\documentclass[twocolumn]{article}

\usepackage[english]{babel}

\usepackage[a4paper,top=2cm,bottom=2cm,left=2cm,right=2cm,marginparwidth=1.75cm]{geometry}

\usepackage{amsmath}

\DeclareMathOperator*{\argmin}{arg\,min}
\usepackage{graphicx}
\usepackage[colorlinks=true, allcolors=red]{hyperref}
\usepackage{amssymb}
\usepackage{gensymb}
\usepackage[square, numbers, super, sort&compress]{natbib}
\usepackage{xpatch}
\makeatletter
\xpatchcmd\citenum{\NAT@parfalse}{\NAT@partrue}{}{}
\makeatother

\usepackage[utf8]{inputenc}
\usepackage{authblk}
\usepackage{svg}
\usepackage{amsmath}
\usepackage{fancyhdr}
\usepackage{tcolorbox}
\DeclareMathOperator*{\median}{median}

\usepackage[font=footnotesize,labelfont=bf]{caption}

\usepackage{colortbl}
\usepackage{tabularx}
\usepackage{xcolor}

\definecolor{lightred}{rgb}{1,0.8,0.8}

\usepackage{multicol}
\usepackage[switch]{lineno}
\usepackage{soul} % For highlighting
\sethlcolor{red!10}

\usepackage{microtype}
\usepackage{placeins}

\tcbset{
    summarybox/.style={
        colback=red!10, % Background color (very light red)
        colframe=red!10, % Border color (dark red)
        fonttitle=\bfseries,
        coltitle=black,
        sharp corners,
        boxrule=0.5mm,
        left=2mm,
        right=2mm,
        top=2mm,
        bottom=2mm,
        width=\linewidth
    }
}

\newcommand{\redline}{\noindent\textcolor{red!70!black}{\rule{\linewidth}{0.5mm}}}

\title{Purely vision-based collective movement of robots}

\author[$2^{**}$]{David Mezey$^{1,2^*}$, Renaud Bastien$^{3}$, Yating Zheng$^{1,2}$, Neal McKee$^{5}$, David Stoll$^{2}$, \\ Heiko Hamann$^{2,4^{**}}$ \& Pawel Romanczuk$^{1,}$}

\affil[1]{Institute for Theoretical Biology, Humboldt University Berlin, Berlin, Germany}
\affil[2]{Science of Intelligence Excellence Cluster, Technical University Berlin, Berlin, Germany}
\affil[3]{Centre de Recherches sur la Cognition Animale, Centre de Biologie Intégrative, \newline Université de Toulouse; CNRS, UPS, France}
\affil[4]{Department of Computer and Information Science, University of Konstanz, Konstanz, Germany}
\affil[5]{Bernstein Center for Computational Neuroscience Berlin, Berlin, Germany}
\affil[*]{Corresponding Author, email: mezeydavid@gmail.com}
\affil[**]{Shared senior authorship}
\date{}

\renewcommand{\headrulewidth}{1pt}
\renewcommand{\footrulewidth}{1pt}
\renewcommand{\headrule}{\hbox to\headwidth{\color{red!70!black}\leaders\hrule height \headrulewidth\hfill}}
\renewcommand{\footrule}{\hbox to\headwidth{\color{red!70!black}\leaders\hrule height \footrulewidth\hfill}}
\begin{document}
\onecolumn
\maketitle
\thispagestyle{fancy}
\redline
\begin{abstract} %181 words
Collective movement inspired by animal groups promises inherited benefits for robot swarms, such as enhanced sensing and efficiency. However, while animals move in groups using only their local senses, robots often obey central control or use direct communication, introducing systemic weaknesses to the swarm. In the hope of addressing such vulnerabilities, developing bio-inspired decentralized swarms has been a major focus in recent decades. Yet, creating robots that move efficiently together using only local sensory information remains an extraordinary challenge. In this work, we present a decentralized, purely vision-based swarm of terrestrial robots. Within this novel framework robots achieve collisionless, polarized motion exclusively through minimal visual interactions, computing everything on board based on their individual camera streams, making central processing or direct communication obsolete. With agent-based simulations, we further show that using this model, even with a strictly limited field of view and within confined spaces, ordered group motion can emerge, while also highlighting key limitations. Our results offer a multitude of practical applications from hybrid societies coordinating collective movement without any common communication protocol, to advanced, decentralized vision-based robot swarms capable of diverse tasks in ever-changing environments.
\end{abstract}
\redline

% One-sentence summary
\section*{\small One-Sentence Summary} %124chars
\tcbset{colback=red!10, colframe=red!70!black}
\begin{tcolorbox}[summarybox]
We present a purely vision-based robot swarm achieving ordered group motion even without central information or communication.
\end{tcolorbox}

\section*{\small Keywords}
\hl{ swarm robotics }, \hl{ visual perception }, \hl{ collective movement }, \hl{ flocking }, \hl{ decentralized design}

%\linenumbers
\twocolumn
\section*{Introduction} %1106 words
%This is a comment
Collective movement is universally present in groups. It is observed in animals across different species and sizes showing fascinating patterns in space and time with remarkable precision---from fish swirling in schools \cite{pavlov2000patterns}, starlings flying in synchronised murmurations \citep{king2012murmurations} or insects marching in organized bands \citep{buhl2006disorder}. Collective movement offers a wide range of benefits for animal groups, such as improved sensing \citep{ioannou2012predatory}, higher foraging efficiency \citep{clark1986evolutionary, pitcher1982fish} or reduced risk of predation \citep{sword2005migratory} among others \citep{beauchamp2021flocking, morelli2019contagious}. In promise of inheriting its advantages in engineered swarms of robots, flocking remains a target behavior of high interest for bio-inspired swarm robotic applications including environmental monitoring \citep{elwin2019distributed, schmickl2011cocoro}, underwater missions \citep{leonard2010coordinated, luvisutto2022robotic}, automation of agricultural tasks \citep{albani2017monitoring, stachniss2022monitoring}, search and rescue in emergency scenarios \citep{arnold2018search}, or even enhanced exploration of the surface of Mars \citep{kang2019marsbee}.

Swarm robotic systems often adopt heuristic rules---designed with a bird's-eye perspective---to foster emergent collective movement in mobile robots. These include aligning with nearby robots, following zonal rules based on distance, or moving towards the group's center of mass.
Even though these strategies might not centralize control over the swarm, they often rely on globally or locally shared information, which is frequently collected, processed or distributed from a central node \citep{yasuda2014self, vasarhelyi2018optimized, cheraghi2020robot, ban2021self, bahaidarah2024swarm}. Examples include distances between robots, relative positions, absolute GPS coordinates, and directional data. Moreover, even when information is gathered locally, it is often communicated between robots through radio, infrared (IR) signals or other channels \citep{turgut2008self, ducatelle2011communication, ferrante2014self, weinstein2018visual, bennis2019short, zhao2020flocking}.
Relying on global knowledge or on direct communication between robots comes with unavoidable costs: firstly, the swarm can solely operate in environments where centrally processed knowledge is reliably accessible at any time, hence becomes vulnerable to tampered or missing information; secondly, the group becomes sensitive to jammed communication channels \citep{higgins2009survey} and to communication delays~\citep{zheng2022experimental} or outages \citep{casas2018impact}. Furthermore, direct information sharing between robots is only possible with compatible communication systems.

Social animals, in contrast to artificial systems, robustly realize collective motion across taxa \citep{moussaid2009collective}. While doing so, they only use locally available sensory information without direct access to positions or heading directions of others.
However, reproducing bio-inspired, decentralized collective motion within robotic swarms has proven to be a complex challenge. Firstly, these challenges arise due to classical bio-inspired flocking algorithms being designed from a global perspective, making central control a straightforward choice~\citep{wiandt2015application}. Secondly, engineering robot hardware with reliable relative localization and communication techniques capable of replacing global information or planning presents its own set of challenges \citep{wang2023distributed}. Furthermore, mapping local interactions to global emergent behavior is a complex task in itself~\citep{hamann2014derivation, hamann2016population, hamann2018swarm}.

The first simple flocking algorithm inspired by social insects that does not rely on communication, memory, or global information was proposed over a decade ago \citep{moeslinger2011minimalist}. This decentralized method, relying exclusively on short-range (local) sensing, is particularly suitable for small robots that have strictly limited capacity in perception, communication, or computation. However, its practical applicability remains limited to small groups.

For larger robots with enhanced capabilities, vision-based flocking has also been suggested as a viable alternative among other sensor-based solutions \citep{chen2022survey}. In such an approach, robots would primarily rely on local visual information, similar to fish \citep{pavlov2000patterns} and birds \citep{bajec2009organized}, where vision is the principal modality supporting collective movement.

To deploy the first autonomous aerial swarm without central control or mutual communication, \citet{petravcek2020bio} utilized a novel relative localization technique~\cite{walter2018mutual} based on ultraviolet markers and light sensors, laying the foundation for vision-based decentralized swarms.

Only recently, \citet{schilling2021vision} presented the first vision-based drone swarm capable of collective motion using only regular onboard RGB cameras. This study shows that with advancements in computer vision, drones can now detect each other during operation without additional visual markers, even in outdoor settings. This approach also demonstrates that purely visual information can drive the collective migration of up to three aerial robots. Although this algorithm relies solely on visual cues, it does so only to estimate the relative distance between drones. This data is fed into a simplified flocking module based on the traditional Reynolds flocking model \citep{reynolds1987flocks}, originally developed for computer graphics. Since estimating relative heading angles and alignment rules from camera images is not trivial, migratory goals were introduced to achieve group motion in common directions. To provide accurate input for the flocking module, the tracking algorithm also uses general assumptions about the environment and the size of other robots. The performance of this algorithm under various sensory or environmental constraints, in scenarios with agents unaware of migratory goals, or in groups larger than three, remains to be evaluated. 

A purely vision-based model of collective motion, first introduced by \citet{bastien2020model}, is a novel concept. This model, based on minimal visual interactions, has been shown to yield collisionless, polarized group motion in simulated agents. By using only low-level sensory information, it fosters an inherently decentralized design and has been suggested as a promising alternative to current flocking controllers in swarm robotics. However, constraints typical for both natural and robotic collectives are yet to be explored. Examples include the effects of a limited field of view and a confined physical environment in which agents are embodied.

Using this simple vision-based model, we demonstrate stable flocking in a swarm of ten differential drive terrestrial robots and lay the theoretical foundation of the model under sensory and environmental limitations using agent-based simulations. While our simulations provide general theoretical insights into collective dynamics in realistic scenarios, our robot experiments validate these findings under the limitations of real-world settings. To our knowledge, this is currently the only robot swarm operating solely based on individual vision, without any centrally computed or explicitly shared information.

By first implementing the model in an agent-based simulation framework, we show that a collisionless and polarized collective motion is achievable, even with a limited field of view (FOV) and a confined arena, that generally reduce group order due to out-of-sight group members. To validate our results, we transfer the model to our purely vision-based robots operating in a confined space. We then show that robots guided by this algorithm can maintain ordered collective movement (1) only using their single, limited FOV camera stream as individual input for flocking, (2) without explicitly estimating the positions or orientations of others, (3) without storing any state information about the environment in memory, (4) without directly communicating their state with others or with any central node, and (5) running all necessary computations onboard.
To conclude our study, we discuss potential future research towards a reliable, purely vision-based swarm of mixed entities.

\section*{Vision-based model of collective motion} % 828 words

\begin{figure*}[t]%[hbt!]
\centering
\includegraphics[width=\textwidth]{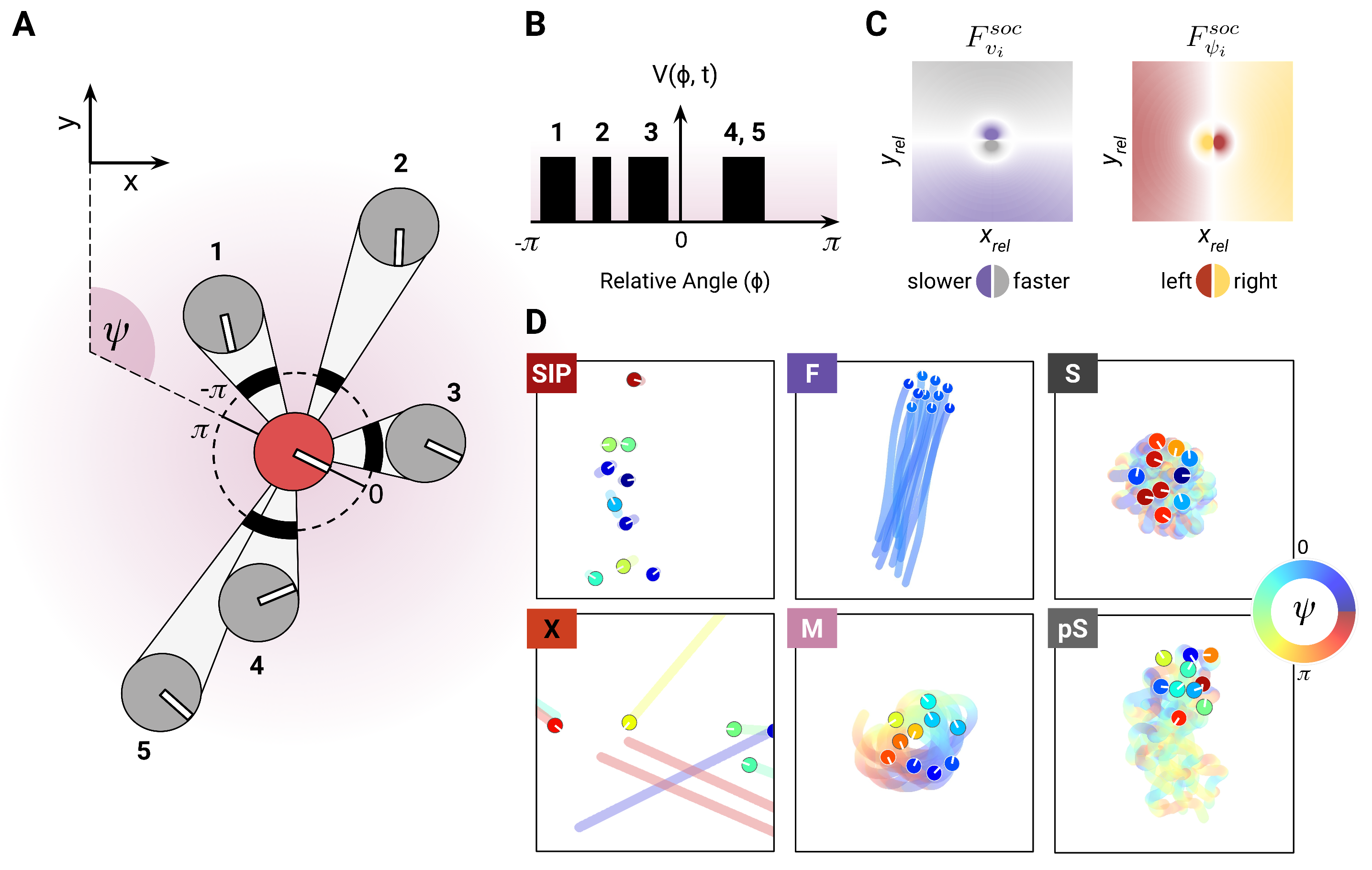}
%\includesvg[inkscapelatex=false,width=1\textwidth]{Figures/VSWRM_figure0_theory.svg}
\caption{\label{fig:fig1} \textbf{Vision-based Model}, \textbf{A}: A focal (red) agent with orientation $\psi$ and with 5 numbered visible neighbors (grey). The focal agent has a full, $2\pi$ FOV (shaded with pink) and an unlimited visual range. The projection of the visual field of the focal agent (represented on a dotted circle) is 1 where other agents are visible (thicker dark arcs), 0 otherwise. \textbf{B}: The resulting unfolded 1-dimensional visual projection field $V(\phi, t)$ of the focal agent. Projection blobs are numbered according to the corresponding visible agents in panel A. When agents partially occlude each other, visual blobs merge (e.g.: agent 4 and 5). \textbf{C}: The front-back (left) and left-right (right) social force fields shaping agent velocity and turning rate respectively. The map depicting $F_{v_i}^{soc}$ displays grey (or purple) at specific relative positions, indicating how a focal agent at the origin would accelerate (or decelerate) in response to the presence of another agent of identical body size at those positions. The map showing \textbf{$F_{v_i}^{soc}$} is grey (or red) at those relative positions where these responses would be turning right (or left). Accurate scales in both maps depend on model parameters with $\alpha_0$ and $\beta_0$ controlling response amplitudes while $\alpha_1$ and $\beta_1$ controlling equilibrium distances from the origin in which no response is given. \textbf{D}: Emergent movement patterns of the vision-based model in toroidal space with full field of view (as in Fig.~\ref{fig:fig_FOV_metrics}). These are \textit{Stuck-in-place} (SIP), \textit{Flocking} (F), \textit{Swarming} (S), \textit{Unordered} (X), \textit{Milling} (M), \textit{Moving swarms} (pS). Colors along the agents' trajectories represent their orientations $\psi$ (See color wheel on the right).}
\end{figure*}

The purely vision-based model of collective motion \citep{bastien2020model} in 2D considers agent $i$ as a disc with diameter $L_0$ moving within a global coordinate system in the direction $\psi_i(t)\in[0,2\pi)$ with velocity $v_i(t)$ (see Fig.~\ref{fig:fig1}A). The model dynamically changes the state of each agent ($v_i(t)$, $\psi_i(t)$) based on its visual input, represented as a visual projection field $V_i(\phi,t)$ (VPF).
This input is a 1-dimensional binary representation of the surrounding visual information with the focal agent itself at the center. It is `1' for relative angles around the focal agent ($\phi_i\in[-\pi, \pi]$) where another agent is visible, and `0' where no other agents are detected. As a result, $V_i(\phi,t)$ acts as a ``visual radar", where every other agent within sight translates into a perceptual representation or visual ``blob" (see Fig.~\ref{fig:fig1}A-B). This binary input of the vision-based model does not distinguish between surrounding peers and the visual representation of multiple agents can overlap, forming larger visual ``blobs" than what individual agents would produce alone (see Fig.~\ref{fig:fig1}B, agent 4 and 5).  In this way, the model's input implicitly accounts for the visual occlusions of visible peers. Furthermore, a strict identification of other agents is not necessary for the resulting applications. This flexibility allows the use of a wide range of detection methods, including LiDAR, IR sensors, CNN-based peer detection, and simple color or other image feature-based segmentation methods.

The state update is described by a set of differential equations using only the visual input and the actual state of the agent (see Eq.~\ref{eq_dvi}-\ref{eq_Fpsisoc} and Sec. \nameref{sup_examples} for further details). Agents move with their preferred speed until others become visible. Upon detecting others, social interactions ($F_{v_i}^{soc}$ and $F_{\psi_i}^{soc}$) are added to the agent's self-propelled movement, adjusting its velocity and turning rate accounting for the presence of visible peers, i.e.

\begin{eqnarray}
    \label{eq_dvi}\frac{\partial {v_i(t)}}{\partial t} &=& F^{ind}(v_i(t)) + F_{v_i}^{soc}(V_i(\phi_i, t)) \\
    \label{dpsi}\frac{\partial {\psi_i(t)}}{\partial t} &=& F_{\psi_i}^{soc}(V_i(\phi_i, t))
\end{eqnarray}

where the first (self-propelled) velocity term in Eq.~\ref{eq_dvi} accounts for agents preference to their individual speed $v_0$ in absence of others, i.e.

\begin{equation}\label{eq_Findv}
    F^{ind}(v_i(t)) = \gamma(v_0-v_i(t))
\end{equation} with parameter $\gamma$ effectively tuning the strength of this preference.

The model generates a social force on each agent emerging from the interplay of attraction and repulsion, influencing both movement velocity ($F_{v_i}^{soc}$) and turning rate ($F_{\psi_i}^{soc}$) as illustrated in Fig.~\ref{fig:fig1}C. The equations governing these social forces read as follows:

%\begin{eqnarray}
%\label{eq_Fvisoc}F_{v_i}^{soc}(V_i(\phi_i, t)) &=& \int_{-\pi}^{\pi} \alpha_0\left\{-V_i(\phi_i, t)+\alpha_1\left(\frac{\partial { V_i(\phi_i, t)}}{\partial \phi_i}\right)^2\right\}\cos\phi_i d\phi_i 
% \\
%\label{eq_Fpsisoc}F_{\psi_i}^{soc}(V_i(\phi_i, t)) &=&
%\int_{-\pi}^{\pi} \beta_0\left\{-V_i(\phi_i, t)+\beta_1\left(\frac{\partial { V_i(\phi_i, t)}}{\partial \phi_i}\right)^2\right\}\sin\phi_i d\phi_i
%\end{eqnarray} 

\begin{align}
\label{eq_Fvisoc}
F_{v_i}^{soc}(V_i(\phi_i, t)) = \int\limits_{-\pi}^{\pi} \alpha_0 A(\phi_i, t) \cos\phi_i d\phi_i
\end{align}

\begin{align}
\label{eq_A}
    A(\phi_i, t) = -V_i(\phi_i, t) + \alpha_1\left(\frac{\partial { V_i(\phi_i, t)}}{\partial \phi_i}\right)^2
\end{align}

\begin{align}
\label{eq_Fpsisoc}
F_{\psi_i}^{soc}(V_i(\phi_i, t)) = \int\limits_{-\pi}^{\pi} \beta_0 B(\phi_i, t) \sin\phi_i d\phi_i
\end{align}

\begin{align}
\label{eq_B}
    B(\phi_i, t) = -V_i(\phi_i, t) + \beta_1\left(\frac{\partial { V_i(\phi_i, t)}}{\partial \phi_i}\right)^2
\end{align}

The total extent of the exerted social force is calculated from the width and position of all visual projection ``blobs", indicating the presence of other agents, by integrating the visual input over the focal agent's retina (see Eq.~\ref{eq_Fvisoc}-\ref{eq_B} and Sec.~\nameref{sup_examples}). The model yields reflexive behavior of agents without relying on any memory of past events, i.e. only the current visual input is considered.

The model has four principal parameters in 2-dimensional space: $\alpha_0$ controls the intensity of acceleration responses to visible others, while $\beta_0$ controls the amplitude of social turning responses. $\alpha_1$ and $\beta_1$ determine the front-back ($L^{fb}_{eq}$) and left-right ($L^{lr}_{eq}$) equilibrium distances between individuals. When a visible peer approaches the focal agent and crosses these equilibrium distances, the overall social force turns from attraction into repulsion in both acceleration and turning respectively.

The visuo-motor system's symmetries, distinguishing front from back and left from right, are modeled through \textit{cos} and \textit{sin} masks across the visual field (see Fig.~\ref{fig:fig_s1} and Fig.~2 in \citenum{bastien2020model}). These masks modulate visual input, enabling direction-specific responses in acceleration and turning respectively. For instance, the \textit{cos} mask changes signs from the front to the rear of the agent (from $\phi_i=0$ towards $\phi_i=\pm\pi$). As a result, the same attraction to a social cue yields positive acceleration when the cue is ahead and negative deceleration when the cue is behind. Note, that in the original model, agents have an omnidirectional FOV, allowing them to perceive others in any direction.

We emphasize that there is no direct alignment rule present in this model. Instead, an effective alignment of agents emerges as a collective behavioral property from the interplay of lower-level local attraction-repulsion interactions, adaptive agent speed, and the symmetries of the agents' visuo-motor systems. Our approach is supported by recent studies showing that explicit alignment rules, while used widely \citep{reynolds1987flocks, vicsek1995novel, couzin2002collective}, are not a prerequisite to achieve polarized collective movement \citep{zheng2022experimental, romanczuk2009collective, ferrante2012self, murakami2017emergence, strombom2022attraction}.

The vision-based model can produce a wide range of collective motion patterns only using simple visual cues (see~\citenum{bastien2020model} and Fig.~\ref{fig:fig1}D). However, it was previously assumed that agents have access to omnidirectional visual information and move in an infinite space---idealized conditions compared to real-world scenarios. 
The effect of typical constraints characteristic to real-world systems, such as a limited FOV and spaces constrained by their walls, which significantly influence collective behavior in zonal models \citep{soria_influence_2019}, was not previously investigated. These constraints apply to most embodied groups including robotic swarms limited by their sensory capabilities and the physical space they move in. Therefore, our work is twofold: 
First, we lay the theoretical foundation by exploring the effects of a limited field of view on the vision-based model and consider the implications of constrained spaces. 
Such investigation is crucial to assess the applicability of the vision-based model for controlling robots;

\FloatBarrier
\begin{figure*}[h!]
\centering
\includegraphics[width=\textwidth]{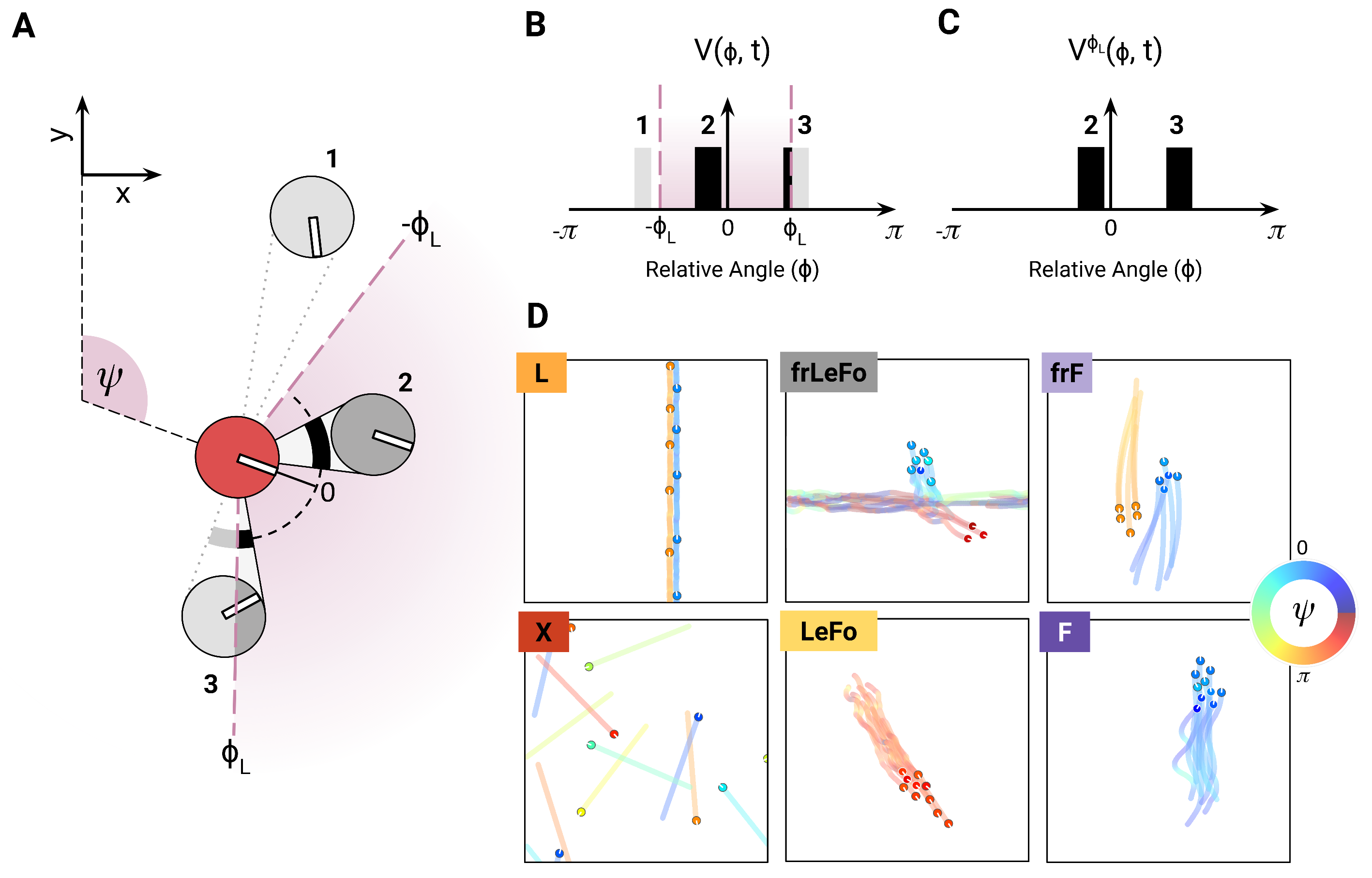}
%\includesvg[inkscapelatex=false,width=1\textwidth]{Figures/VSWRM_figure1_fov_patterns.svg}
\caption{\label{fig:fig_FOV_patterns} \textbf{Limiting the active FOV}, \textbf{A}:  A focal (red) agent with orientation $\psi$, with a limited FOV between relative angles $[-\phi_{L}, \phi_L]$ highlighted with pink between pink dotted lines and with three other agents (grey), either fully visible (2), partially visible (3) or not visible (1). In-sight parts of other agents are colored darker grey. \textbf{B}: The original 1-dimensional visual projection field $V(\phi, t)$ with black visual blobs for in-sight, and light grey for out-of-sight visual information. The active field of view is highlighted with pink between pink dotted lines. \textbf{C}: The resulting 1-dimensional limited projection field $V_i^{\phi_L}(\phi, t)$. Partial visual blobs (3) are fully recovered to avoid boundary effects coming from a limited FOV (See Sec. \nameref{sup_limfov}). \textbf{D}: Emergent movement patterns (As in Fig. \ref{fig:fig_FOV_metrics}) of $N_A=10$ agents on a torus arena and with limited FOV. These are \textit{Lines} (L), \textit{fragmented Leader-Follower} (frLeFo), \textit{fragmented Flocking} (frF), \textit{Unordered} (X), \textit{Leader-Follower} (LeFo), \textit{Flocking} (F). Colors along the agents' trajectories represent their orientations $\psi$ (See color wheel on the right).}
\end{figure*}
\FloatBarrier

Second, we implement the model as a robot controller and transfer it to ten terrestrial robots forming a vision-based swarm. Our approach is truly decentralized from robot perception to every computational step executed onboard. We then demonstrate the collective movement of this swarm in real-world conditions, explore the effect of key model parameters on movement order, and showcase some of the primary benefits of the vision-based swarm through four exemplary videos\cite{mezey2024video4k}.

\subsection*{Quantifying Collective Behavior}\label{quantcollbeh} %298 words
During our analyses, we focus on five key metrics described in detail in Sec. \nameref{sup_metrics}. Here we give a short overview of these: We quantified group polarization ($P$), mean inter-individual distance ($\tilde{D}$), the overlap or collision avoidance time ratio ($R_o^{sim}$, $R_o^{exp}$), group circularity ($RCA$), and the number of largest subgroup ($N_{clus}^{max}$) after hierarchical clustering of the agents to quantify group fragmentation. 
In short, polarization ($P$) indicates the uniformity of movement direction within the group, achieving high values close to 1 when all agents move in similar directions while low values close to zero for unorganized movements where agents or subgroups move in different directions.
$R_o^{sim}$ in simulated agents shows what proportion of simulation time (in percent) agents spent overlapping and is directly related to agent-agent collision probability in embodied groups. As overlaps are not possible in a physical space, during robot experiments $R_o^{exp}$ shows what proportion of experiment time (in percent) robots spent with avoiding collisions with each other (see Sec. \nameref{sup_collisions}). $R_o^{sim}$ reaches 100 if all agents overlapped through the whole simulation time and 0 if no agent-agent overlaps were observed. Similarly, $R_o^{exp}$ reaches 100 if every robot was in collision avoidance during the whole experiment time, and 0 if no collision avoidance event has been triggered. The circularity of groups ($RCA$) describes the shape of the group. An $RCA$ value of 1 signifies a perfectly circular group, whereas values approaching 0 indicate elongated formations. Furthermore, we clustered agents according to their movement similarity (i.e. similarity based on heading angle and relative distance) and measured the size of the largest cohesive polarized subgroup ($N_{clus}^{max}$) to quantify group fragmentation (for further details see Sec. \nameref{sup_metrics}). For fully fragmented groups this value is one, while in fully cohesive group this value reaches the number of agents in the group ($N_A=10$).

\subsection*{Field of View}\label{fov}
\subsubsection*{Model Modifications}\label{fov_model} % 375 words
We started our investigation by implementing the vision-based model in an agent-based simulation framework \citep{mezey_P34ABM_A_novel}. In the first set of simulations, $N_A=10$ disc-shaped agents were uniformly distributed across a toroidal arena, i.e. across an arena with periodic boundaries on opposite sides (see also Sec. \nameref{sup_torus}). The width and height of the arena were set to 900~pixels, which is approximately 164 times the radius of agents ($R_A=5.5$ pixels) similar to experimental conditions (see Sec. \nameref{robot_exp}). The agents' FOV varied from 0 (equivalent to blind agents) to $2\pi$ (representing unobstructed vision), allowing us to systematically explore the impact of FOV restrictions on the vision-based collective behavior.

The choice of a toroidal arena serves a dual purpose: (1) It allows us to directly compare our results with prior studies that assume agents operate in infinite space with unrestricted FOV, while (2) preventing catastrophic group fragmentation---a possible issue in infinite environments where agents with limited FOV could permanently lose visual contact with each other. The periodic boundaries circumvent this problem by ensuring that individuals exiting one side of the arena reappear on the opposite side, sustaining visual contact and group integrity even with a limited FOV.
 
The agents' FOV was restricted by setting their visual projection fields $V_i(\phi, t)$ to zero for all relative angles $\phi$ exceeding the FOV limits $\pm\phi_L$, thereby creating a limited visual field $V_i^{\phi_L}(\phi, t)$. If another agent is only partially visible at the peripheries of the active FOV, the entire visual projection blob corresponding to that agent is reconstructed (see Sec. \nameref{sup_limfov} and Fig.\ref{fig:fig_FOV_patterns}A-C). Limiting the FOV this way effectively introduces a blind spot behind each agent, such that they perceive information only around the primary direction of movement (front).  

By default, visual range of agents is technically unlimited but only visual blobs wider than a single retinal pixel are considered during the composition of the visual projection fields. This way the resolution of the projection field ($N_{ret}$) ultimately determines the maximum range of vision similarly to actual visual sensors. Lower resolutions result in coarser visual fields, which may fail to detect distant agents if their visual representation is too small. The resolution of the simulated agents' retinas is aligned with what is achievable on our robot platform (see Table \ref{tab_simulationparams} and \ref{tab_experiment}).

\subsubsection*{Results}\label{res_fov} % 895 words
To study the influence of a limited FOV on the collective movement of simulated agents, we conducted a series of simulations where we varied the model's parameters $\alpha_0$, $\beta_0$ and the agents' FOV, while other parameters remained constant as detailed in Table~\ref{tab_simulationparams}. These simulations were repeated 20 times, exploring FOVs from 25\% to 100\% of the complete range of $2\pi$. We quantified the collective behavior through key metrics described in Sec. \nameref{quantcollbeh}.

\begin{figure*}[hbt!]
\centering
\includegraphics[width=0.9\textwidth]{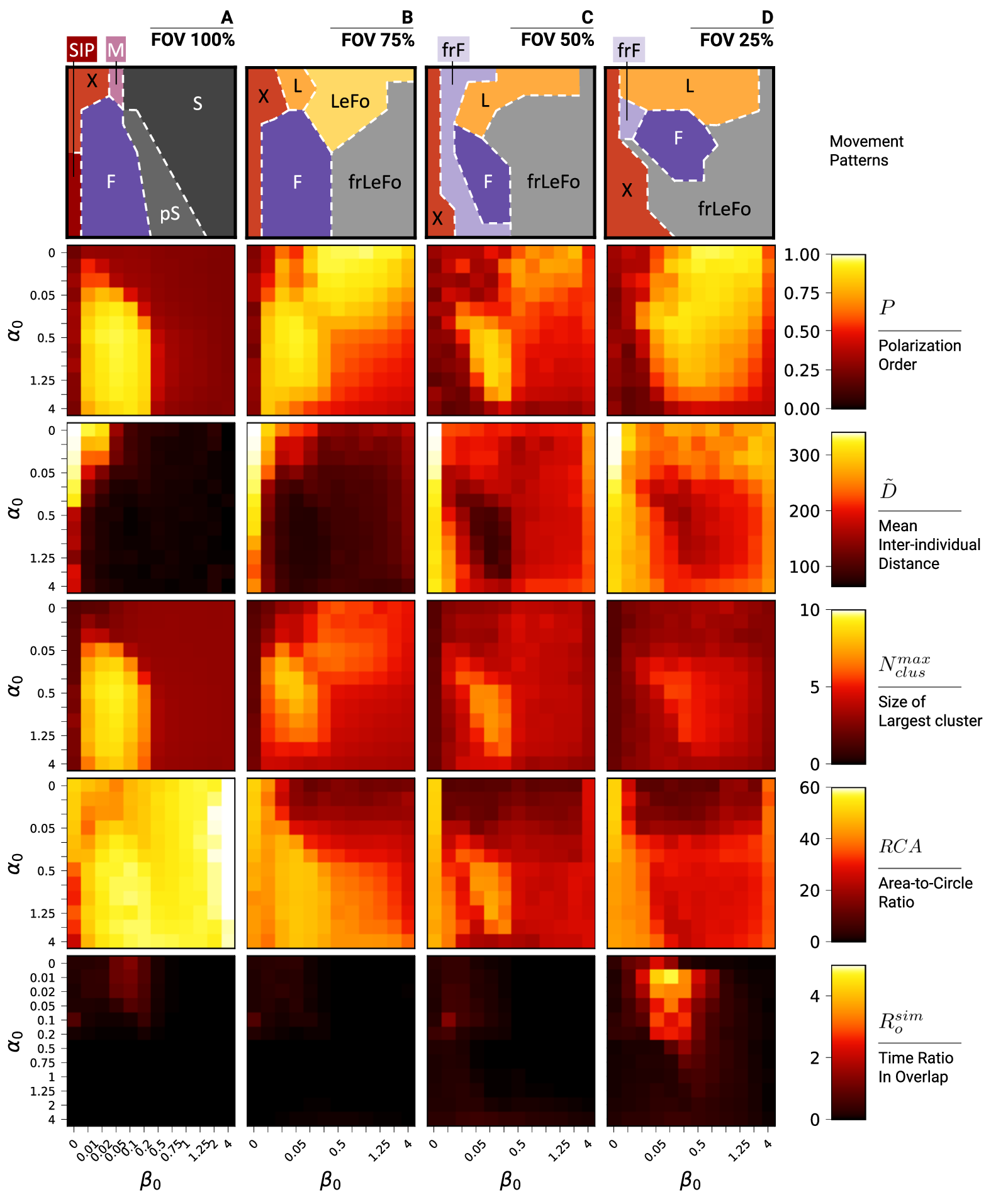}
%\includesvg[inkscapelatex=false,width=0.9\textwidth]{Figures/VSWRM_figure1_fov_metrics.svg}
\caption{\label{fig:fig_FOV_metrics} \textbf{Effects of a limited FOV on a torus.} Emergent collective movement patterns as depicted in Fig.~\ref{fig:fig_FOV_patterns} (top row) and summary metrics (rows) of the resulting collective movement with different fields of view (columns) over different $\alpha_0$ (y axis) and $\beta_0$ (x axis) parameters. Simulations with full FOV align with the results of \citet{bastien2020model}. Observed movement patterns are \textit{Unordered} (X), \textit{Leader-Follower} (LeFo), \textit{fragmented Leader-Follower} (frLeFo), \textit{Flocking} (F), \textit{fragmented Flocking} (frF), \textit{Lines} (L). Reducing agents' FOV facilitates polarized movement via leader-follower (LeFo, frLeFo) dynamics. Due to a blind spot behind agents, groups with lower FOVs are less cohesive than those with a full field of view. Collisionless polarized flocking behavior was observed for all FOV values larger than 25\%.}
\end{figure*}

Our findings, aligning with \citet{bastien2020model}, show that groups with unobstructed FOV exhibit behaviors from flocking, with high group polarization and cohesion, to milling and swarming under varied $\alpha_0$ and $\beta_0$ conditions (see Fig.~\ref{fig:fig_FOV_metrics} column A). Flocking (F) is observed where agents can sufficiently modulate their speed ($\alpha_0$ is large enough) but they are not overly maneuverable ($\beta_0$ is relatively small), leading to cohesive and circular group formations with minimal overlaps (Fig. \ref{fig:fig_FOV_metrics} column A, pattern F). 

Lower speed modulation ($\alpha_0$) gives rise to either milling behavior (M) with small $\beta_0$ conditions or swarming (pS, S) behavior with large $\beta_0$ conditions.
During milling agents follow circular paths around the group's center of mass. Swarming is characterized by continuous turning in response to visible agents. This can lead to cohesive movement as a swarm (pS) or to the fixation of the group's center of mass in a stationary position (S). 
When $\beta_0$ is too small agent's do not turn sufficiently in the presence of others to implement organized collective motion, hence with low speed modulation ($\alpha_0$) they are unorganized (X), or when speed modulation is high enough they remain stuck in place (SIP).

Reducing the FOV to 75\% (seeFig.~\ref{fig:fig_FOV_metrics}, column B) reveals a new parameter range with high group polarization at low $\alpha_0$ and high $\beta_0$.
In such settings, agents with a full FOV typically exhibit swarming behavior (S). With a limited FOV in this regime, polarized collective movement can still emerge: as agents with restricted FOVs may break away from the group when they lose sight of others due to the blind spot behind them, others will follow the ``blind" leader causing an overall polarized movement of the group. This state of polarized movement, resulting from the underlying leader-follower dynamics (LeFo), shows lower circularity than flocking behavior. It also tends to oscillate between polarized and non-polarized states, reflecting the remains of true swarming states (S) in case of a full FOV. 
Similarly, when both turning and speed modulation are high, swarming states (pS, S) at a full FOV break down into intermittent polarized and depolarized states. As the group begins to swarm it becomes fragmented as one or more agents break away. These newly emerged "blind" leaders are soon followed by others, leading to a fragmented leader-follower pattern (frLeFo). In a periodic environment, such as a torus, subgroups eventually merge, sustaining an endless cycle of fission and fusion. Furthermore, we observed an elongated movement pattern (L) due to a combination of the periodic environment and the limited FOV, for both low $\alpha_0$ and $\beta_0$ conditions. These patterns show agents in single or multiple lines moving in the same (fully polarized) or opposite directions (anti-polarized), and they were only occasionally observed with a 75\% FOV.

Reducing the FOV further to 50\%  (see Fig.\ref{fig:fig_FOV_metrics} column C) decreases polarization for two main reasons. Firstly, a narrower active visual field increases the likelihood of agents losing sight of each other, resulting in the group splitting into smaller subgroups. While these subgroups might be polarized within themselves, their diverging movement directions diminish overall group polarization. 
Secondly, a narrow field of view complicates rejoining into a single polarized group when the group becomes fragmented. This challenge arises because agents lack visual information about subgroups, some typically being ahead while others behind. Consequently, with lower~$\beta_0$ values, we observe periodic fragmentation into flocking subgroups that attract each other from opposite directions but pass through each other without being able to merge into a single cohesive group (frF). We noted elongated polarized or antipolarized lines (L) appearing more frequently and across a broader range of model parameters at a 50\% FOV than a 75\% FOV. Such movement patterns coexisted with other, usually leader-follower patterns.

Further reducing the FOV to 25\% (see Fig.~\ref{fig:fig_FOV_metrics} column D) intensifies the challenges to group cohesion seen with a 50\% FOV. Agents break away easier from others and rejoining to a cohesive group is difficult as agents are deprived of most of their visual input. The narrow frontal field of view further stabilizes elongated line patterns (L).

In sum, limiting agents' active visual field, while fostering group polarization through temporary ``blind" leaders, poses fundamental challenges to group cohesion. The larger the blind spot, the easier it is for agents to lose track of each other eventually leading to group fragmentation. Subgroups with limited FOV often show difficulty in rejoining to a single cohesive group after fragmentation. Nonetheless, with periodic boundary conditions, we observed a part of the parameter space where polarized cohesive flocking motion persisted for all FOVs. Our results highlight the resilience of the vision-based model and the resulting collective behavior despite significant visual limitations and suggest a wide applicability in the design of robotic swarms even in environments where sensory input is restricted.

\subsection*{Confined Spaces}\label{walls}
\subsubsection*{Model Modifications}\label{walls_model} % 171 words
In realistic scenarios, agents often have to navigate in structured environments constrained by their physical boundaries. Repeatedly interacting with walls might fundamentally change collective behavior, potentially preventing movement coordination or facilitating transitions between behavioral states reported in fish, computational models, and drone swarms \citep{hemelrijk_emergence_2010, tunstrom_collective_2013, vasarhelyi2018optimized}.
To study the influence of space confinement on the vision-based model, we implemented reflective boundary conditions in the simulations. When agents reach these boundaries they must turn back to the arena orthogonal to their incidence angles, i.e. their new heading angle $\psi_i(t+1)$ is chosen as
\begin{align}
    \psi_i(t+1) &\in \left\{ (\psi_i(t) \pm \frac{\pi}{2}) \mod 2\pi \right\}
\end{align}
such that:
\begin{align}
    \quad x_i(t+1) &\in [x_{\min}, x_{\max}] \quad \text{where}\\
    \quad x_i(t+1) &= x_i(t) + v_i(t) \cos(\psi_i(t+1)) 
\end{align} and
\begin{align}
    \quad y_i(t+1) &\in [y_{\min}, y_{\max}]  \quad \text{where}\\
    \quad  y_i(t+1) &= y_i(t) + v_i(t) \sin(\psi_i(t+1))
\end{align} where $v_i(t)$ is the velocity of agent $i$ in time $t$ at position $(x_i(t), y_i(t))$ with heading angle $\psi_i(t)$ and vertical walls are at $x_{min}$ and $x_{max}$, while horizontal walls are at $y_{min}$ and $y_{max}$.

Interactions with boundaries precede any vision-based behavior, meaning agents reflect off walls even if their vision-based model suggests otherwise. We assume agents can detect walls at short range, which can be implemented for example via IR sensors (see Sec. \nameref{sup_collisions}).

\begin{figure*}[hbt!]%[hbt!]
\centering
\includegraphics[width=0.9\textwidth]{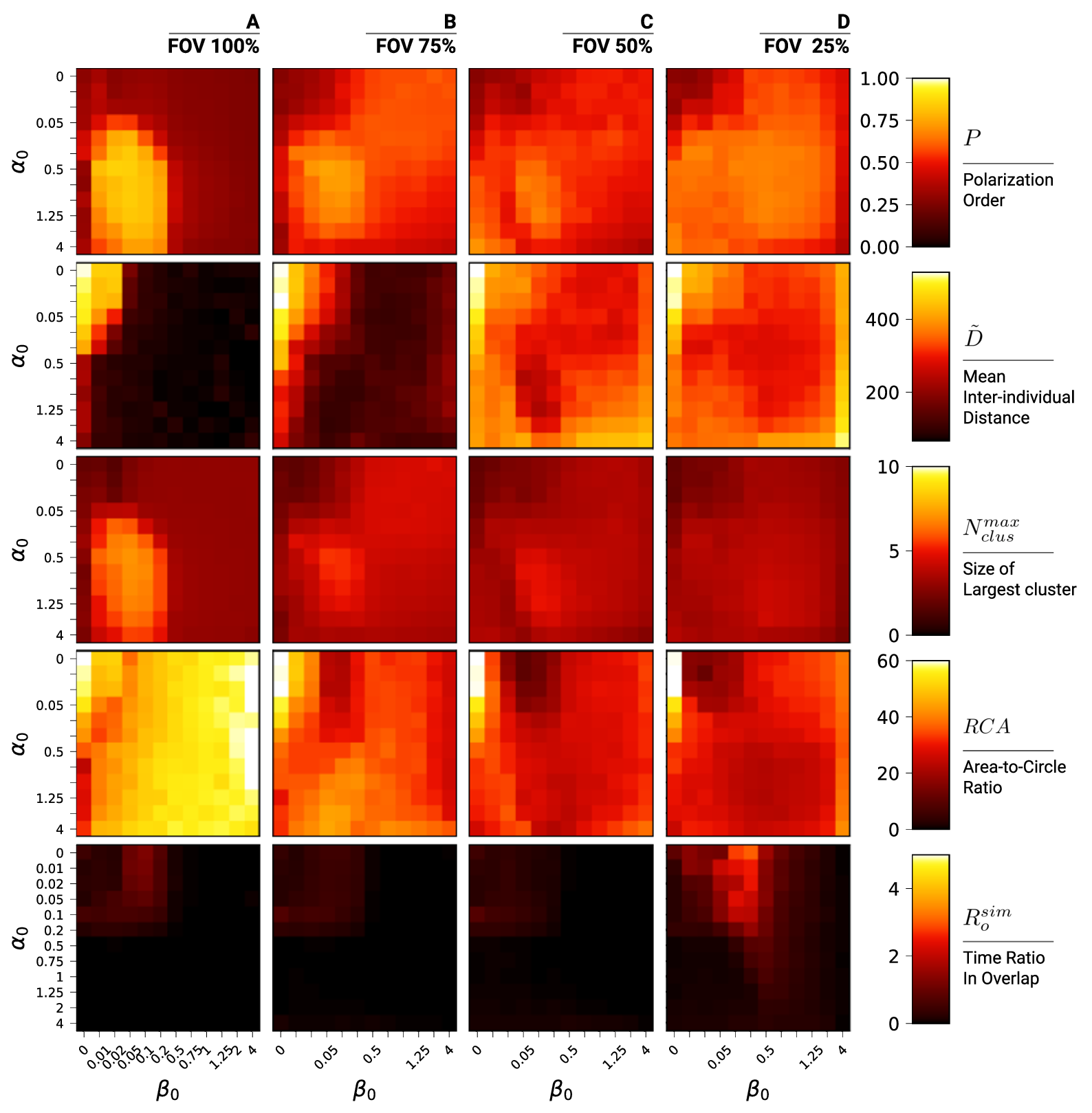}
%\includesvg[inkscapelatex=false,width=0.9\textwidth]{Figures/VSWRM_figure2_wall_metrics.svg}
%\includegraphics[width=\textwidth]{VSWRM_figure2_wall_metrics.eps}
\caption{\label{fig:fig_wall_metrics} \textbf{Effects of a confined arena}, Summary metrics (rows) of the vision-based collective movement with different FOVs (columns) over different $\alpha_0$ (y axis) and $\beta_0$ (x axis) parameters in a confined arena with reflective walls. Introducing boundaries generally decreases polarization and cohesion of the groups and makes it challenging to identify stable movement patterns.}
\end{figure*}

\subsubsection*{Results}\label{res_walls} % 765 words
We simulated the movement of $N_A=10$ agents with the same parameters as in the previous section (see Table~\ref{tab_simulationparams}), and systematically compared different environmental conditions to set the stage for experimentation with robots in real-world environments. We described the resulting collective movement through the same summary metrics described in Sec.~\nameref{quantcollbeh}. The introduction of walls imposed significant and regular perturbation on movement, making it challenging to identify stable movement patterns, such as flocking, milling or swarming states. 
Instead, we identified a specific set of parameters that produced collective movement which was comparatively cohesive, collisionless, and highly polarized, even within reflective walls. This suggests the possibility of a stable flocking state and holds promise for developing controllers for embodied agents.

In confined spaces, agents with full FOV remain cohesive and, after a transition period, resume their original movement patterns despite the perturbation caused by the walls. Due to these temporary unorganized states, polarized groups in a limited space show on average lower polarization, less cohesion and lower circularity than those on a torus, especially with lower FOVs. On the other hand, the introduction of parallel walls causes agents to move in similar directions once reflected. This increases the polarization of otherwise unorganized groups with low polarization. These groups, which cannot modulate their turning sufficiently without walls (i.e., with low $\beta_0$), show higher polarization than those in toroidal arenas.

When agents' FOV is limited, walls become a significant disruptive factor on both group polarization and group cohesion. By repeatedly perturbing the group, walls can suddenly change agents visual input depending on which direction they turn back towards the arena. As a result, agents might easily break away from the group due to their lack of memory of past positions and their blind spot.

In conditions where leader-follower dynamics resulted in stable polarized movement on a torus (see Fig.~\ref{fig:fig_FOV_metrics}, LeFo, frLeFo), walls fully deteriorate polarization, as these states are particularly fragile to external perturbations. If a temporarily ``blind" leader is forced by a wall to turn back towards the group that follows it, the group loses polarization until a new leader can emerge. This effect was consistent across all parameters for limited FOVs.

Walls also reduce group cohesion resulting in higher inter-individual distances and less circular group formations than in toroidal environments. In terms of collisions, groups in confined spaces performed similarly to those on a torus if both control parameters $\alpha_0$ and $\beta_0$ were large enough, such that agents were both sensitive and maneuverable enough to avoid dangerously short distances with others.

The parameter regime yielding the most polarized movement was consistent with those corresponding to flocking states (F) in periodic environments, although polarization notably decreases when reflective boundaries are introduced. These ``islands" of polarized states show comparably smaller inter-individual distances and elongation than other parameter combinations suggesting more cohesive groups. Also, with FOVs higher than 25\%, we observed close to zero overlap ratios in at least parts of these regimes, indicating effective collision avoidance. Due to these considerations they are the best candidates to use for controllers realizing group migration in robotic applications. 

Our results show that confining the arena with reflective walls reduces group cohesion with limited FOVs. Reflections from walls can lead agents to lose sight of each other, promoting group fragmentation---a fundamental challenge in groups with limited perceptual capabilities moving in confined spaces, without the possibility of relying on memory or an explicit model of the surrounding environment. The interplay between the effects of sensory and environmental constraints we presented clearly shows the complexity of achieving effective collective movement in real-world scenarios, and the necessity of further exploration through embodied systems such as a physical robot platform.

\FloatBarrier
\begin{figure*}[h!]
\centering
\includegraphics[width=\textwidth]{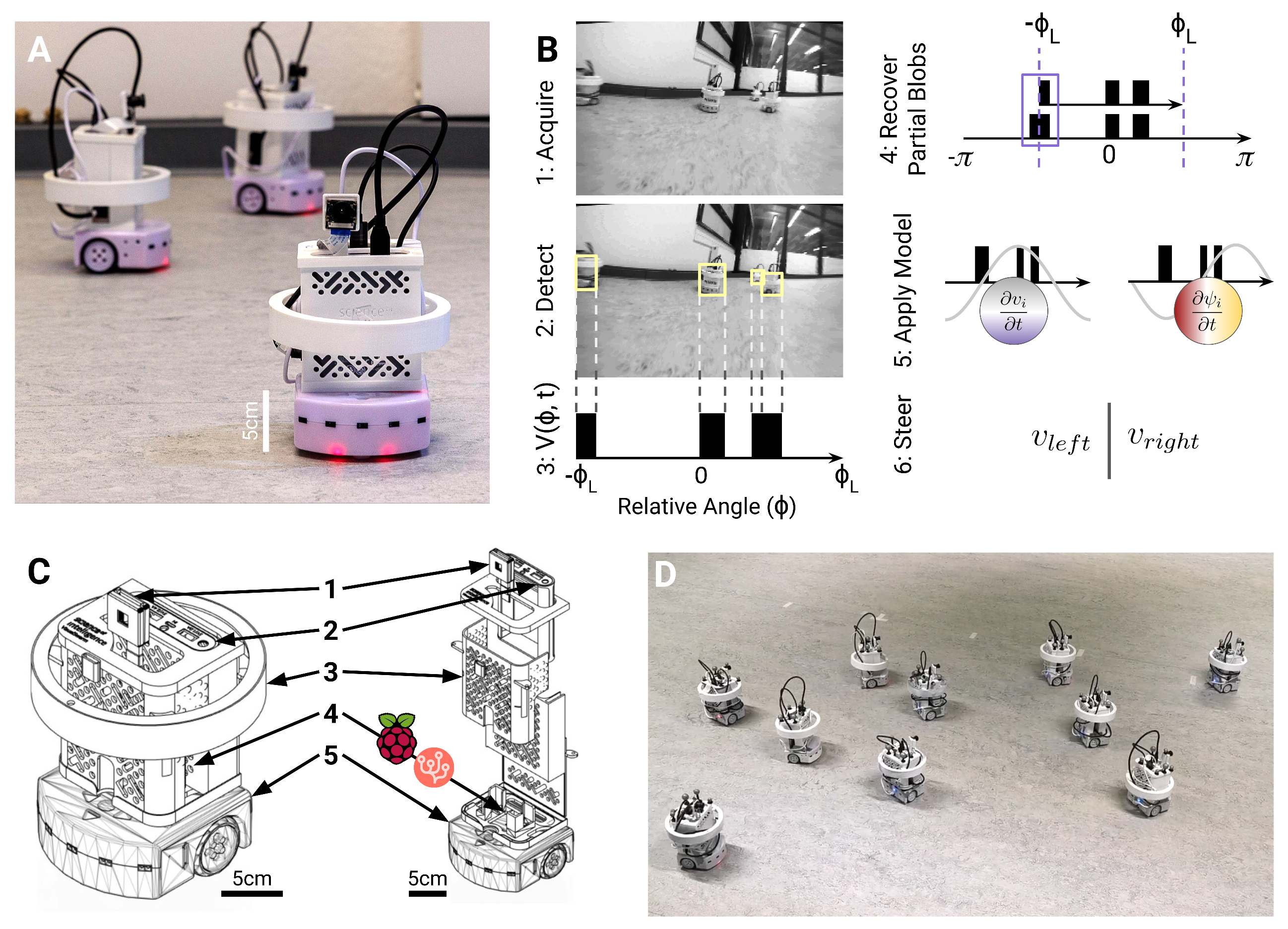}
%\includesvg[inkscapelatex=false,width=1\textwidth]{Figures/VSWRM_figure3_robotplatform_withdesign.svg}
\caption{\label{fig:fig_robotplatform} \textbf{Robot Platform}, \textbf{A}: Close up image of three vision-based robots. \textbf{B}: Computational steps of the robot controller implementing vision-based collective movement. 1.: A camera image is acquired, and unwarped from fisheye lense distortion. 2.: Other robots are detected on the image on board using a tensor processor unit (TPU). 3.: Detection boxes (yellow) are projected to the horizontal axis to create the limited projection field. 4.: Partial projection blobs are recovered using the height of the detection box to create the full visual projection field (VPF). 5.: The VPF is fed into the vision-based model to calculate the desired robot state. 6.: Motor commands are calculated and executed by the robot. \textbf{C}: Hardware components of a vision-based robot: 1: Camera module, 2: Battery, 3: 3D-printed structural scaffolding, 4: Socket for Raspberry Pi 4B and Google Coral EdgeTPU, 5: Thymio~II base robot. \textbf{D:} Swarm of ten vision-based robots}.
\end{figure*}
\FloatBarrier

\hyphenation{Vision-based decentralized flocking of robot}
\section*{\raggedright Vision-based decentralized flocking of robots}
\subsection*{Robot Platform} % 625 words
After exploring the effects of realistic limitations on the vision-based model, we transferred it to terrestrial robots through a controller software~\citep{mezeyVSWRM}. We use the Thymio~II~\citep{mondada_bringing_2017} educational robot platform extended with a Raspberry Pi 4B development board and an RGB fisheye camera module, having a roughly half horizontal FOV (approximately 175°). Robots perceive their environment via the camera stream and detect other robots in the registered image using a convolutional neural network (CNN) based object detector (see Sec. \nameref{sup_detectiontomovement} and Fig.~\ref{fig:fig_robotplatform}B, steps 1-2). Detection runs onboard using a Google Coral tensor processing unit (TPU), generating a 1D binary visual projection field (VPF) as the input for the controller algorithm (See Fig.~\ref{fig:fig_robotplatform}B, step~3). If other robots are only partially visible on the peripheries of the camera image, the height of the detection box is used to recover the entire projection "blob" in the resulting projection field. This is done similarly to our "blob"-recovery method during simulations (see Fig.\ref{fig:fig_FOV_patterns}, B-C) to avoid boundary effects caused by a limited FOV (see Sec. \nameref{sup_limfov} and Fig.\ref{fig:fig_robotplatform}B, step 4). Detection of other robots runs at 5~Hz on an input resolution of 320~$\times$~200 pixels. Note that any method capable of selectively detecting other robots around the focal one would suffice to generate the visual projection field, e.g., other selective segmentation techniques or LiDar-based detection methods~\citep{yilmaz2022lidar, dietrich2019deep}. We used CNN-based object detection due to its widely explored applicability, relatively low price, the recent improvements in inference speed on edge-devices, and its relatively high reliability in changing light conditions.

\begin{figure*}[p]
\centering
\includegraphics[width=\textwidth]{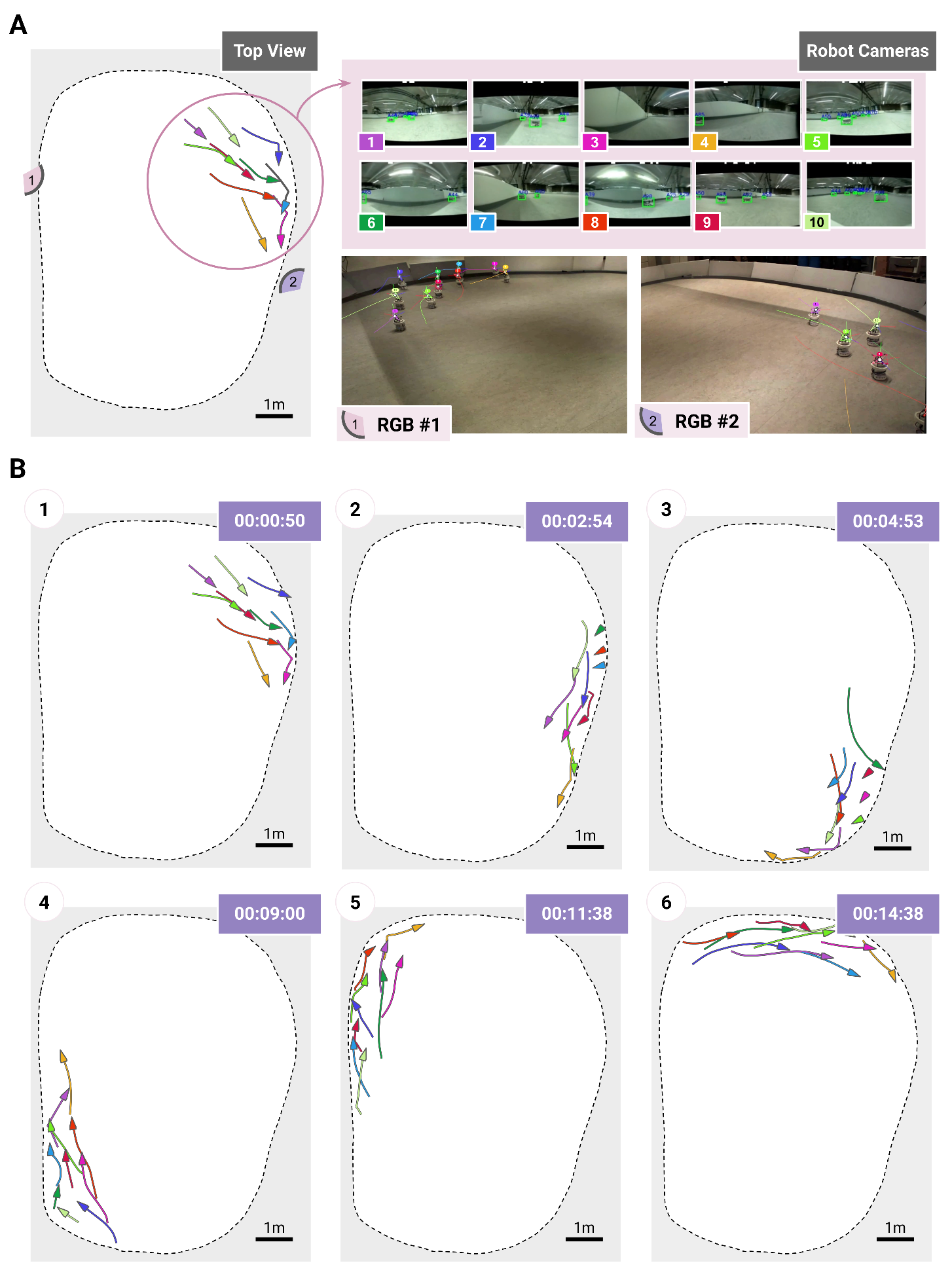}
%\includesvg[inkscapelatex=false,width=1\textwidth]{Figures/VSWRM_figure5_experimentreel.svg}
\caption{\label{fig:fig_robotexpreel} \textbf{Experimental setup for vision-based robot flocking} \textbf{A} Example data sources from a robot experiment. These are as follows: (1) The top view reconstruction of tracked robot positions and orientations (left), with the arena wall depicted with dotted lines and the positions of the two RGB cameras shown with colorful arcs. Reconstructed robots are color-coded according to their IDs. (2) The onboard RGB camera streams of individual robots with detection boxes (top right). The individual camera streams are color-coded according to the robot IDs in the top view reconstruction. (3) RGB camera images with tracking overlay (bottom right). The positions and angles of the RGB cameras are shown on the top view reconstruction with corresponding pink (RGB~\#1) and purple (RGB~\#2) arcs. \textbf{B} Reconstructed top view frames of an experiment showing stable flocking behavior of ten robots. The frame order from one to six is in the top left, and the corresponding experiment time is in the top right corner of each chosen frame. The same experiment is presented in Video~1~\cite{mezey2024video1}}
\end{figure*}

After calculating the visual input for the controller, velocity and turning rate values are determined according to the vision-based model. These are then translated into motor commands for the robot's 2-wheel differential motor system, resulting in movement (see Fig.~\ref{fig:fig_robotplatform}B, steps 5-6). To guard the integrity of the robots in experimental scenarios we used the seven built-in IR proximity sensors of the Thymio~II robot platform. We used these sensors to (1)~avoid possible robot-robot collisions and (2)~to turn robots away from walls of the arena similarly to wall-reflections in simulation experiments. When close to an object (whether a wall or another robot), the robot turns away from the object until the path ahead is clear, irrespective of the visual input (see Sec. \nameref{sup_collisions}).

We built a swarm of ten robots (see Fig.~\ref{fig:fig_robotplatform}D). For our analysis, we collected the position and orientation data of the swarm using an OptiTrack motion capture system with 20 infrared tracking cameras, which detect reflecting markers attached to the robots~\citep{furtado2019comparative}. Additionally, we collected individual camera streams from the robots, published along with calculated detection boxes and the resulting projection fields via the local network, as well as two external RGB streams covering the whole arena (see Fig.~\ref{fig:fig_robotexpreel}A and Video~1\cite{mezey2024video1}). The collected tracking data was used solely for later analysis and was not distributed to the robots during experiments.

We conducted all robot experiments in a 9m$\times$6m arena with flexible grey plastic walls of 8cm height. To minimize false visual detections of external objects outside the arena, we installed an additional wall with a height of 40cm. Being approximately twice the height of the robots, this wall blocks the robots' view once they are near the arena edge.

Robots were initialized in a polarized configuration centered around a randomly selected position near the arena wall. They were started and stopped sequentially from a central computer but did not receive any information from this central node or explicitly communicate their internal states with each other during experiments. In other words, robots operated in a fully decentralized manner. They acquired local visual information using their camera stream, detected other robots using onboard detection, and calculated and executed resulting motor commands onboard as well.

\subsection*{Mobile Robot Experiments}\label{robot_exp} %1353 words

Based on our simulated groups, we chose a range of model parameters where cohesive, polarized motion can emerge. We varied the two main parameters in this parameter range and conducted two sets of about 60-minute recordings for each combination. By that we produced more than 30 hours of experimental data of collective movement of ten robots~\cite{mezey2024robotdata}. We either fixed the maneuverability parameter ($\beta_0$) while adjusting the social acceleration value ($\alpha_0$), or vice versa. As before, we quantified the resulting collective movement through the metrics described in Sec. \nameref{quantcollbeh}. We filtered the recorded data by only including data points where none of the robots were executing collision avoidance maneuvers due to walls, i.e. where all robots acted according to their vision-based controller.

Supporting our results with simulations, we found that polarized cohesive motion stably arise in our robot swarm on a wide range of model parameters. Polarization of the group was highest when both $\alpha_0$ and $\beta_0$ were large enough, allowing robots to sufficiently modulate both their velocity and turning rate in response to visible others. In these conditions robot collision events were unlikely and the group remained polarized for long timescales aligning with the walls of the arena. Group fragmentation was possible due to a limited FOV but rare.

\begin{figure*}[h!]
\centering
\includegraphics[width=\textwidth]{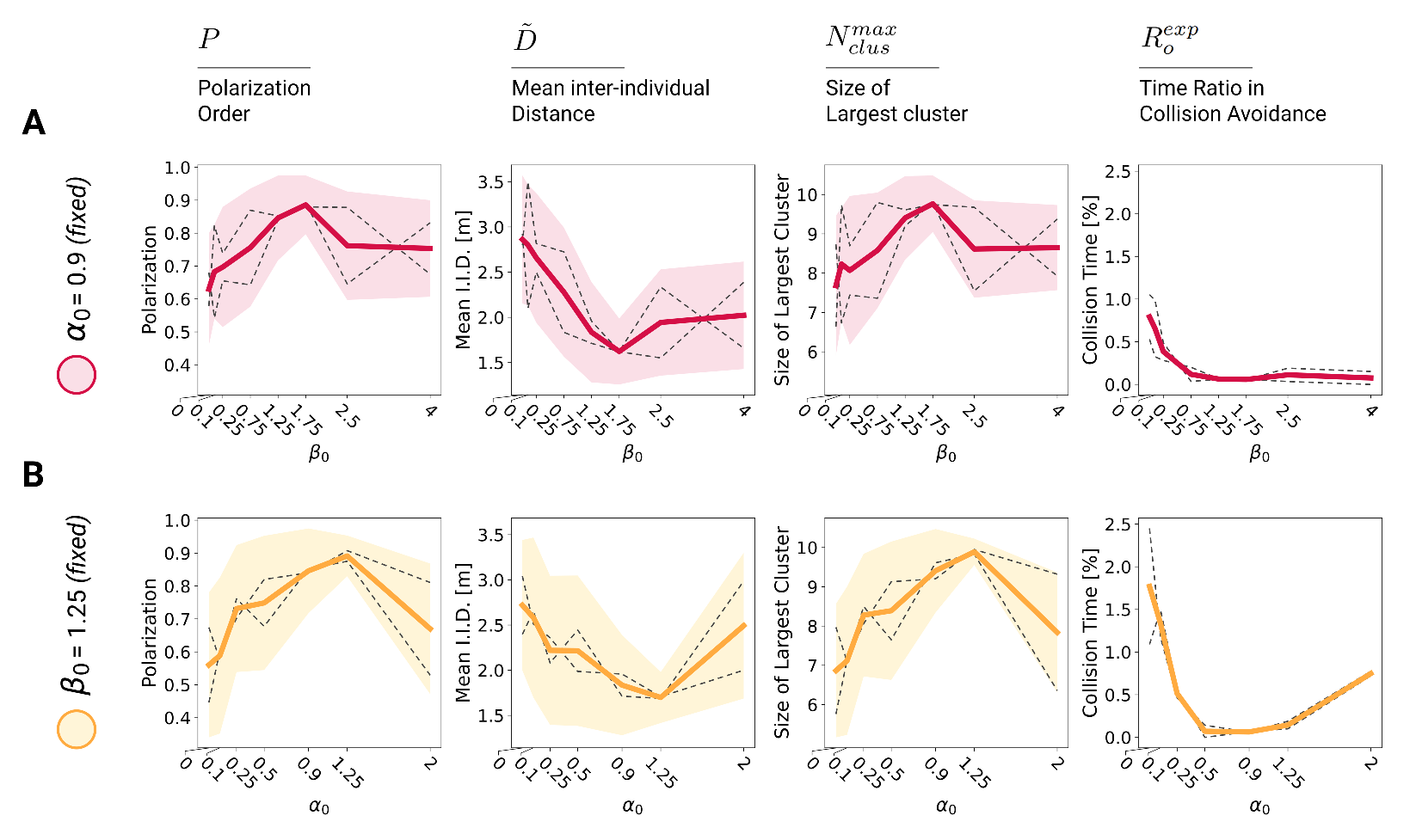}
%\includesvg[inkscapelatex=false,width=1\textwidth]{Figures/VSWRM_figure4_robotexperiments_hor.svg}
\caption{\label{fig:fig_robot_exp} \textbf{Effect of model parameters on robot flocking.} \textbf{A:} The effect of varying $\beta_0$ (x-axes) with fixed $\alpha_0=0.9$ (red) on summary metrics (columns), i.e. group polarization ($P$), mean distance between robots ($\tilde{D}$), size of the largest cohesive, polarized subgroup ($N_{clus}^{max}$), and time ratio spent with robot collision avoidance ($R_o^{exp}$). \textbf{B:} The effect of varying $\alpha_0$ (x-axes) with fixed $\beta_0=1.25$ (orange) on the same metrics from left to right respectively. Thick colored lines in both panel A and B depict temporal means over two repetitions of 60 minutes, dotted grey lines the means of individual repetitions, shaded areas are the standard deviations (STD) over time. Collision avoidance events are binary, hence STD is not provided in this case. In both cases, either when $\alpha_0$ or $\beta_0$ was fixed, robots were able to (1) maximize group polarization and (2) cohesion and (3) minimize collisions with some mid-range optimal parameter combinations.}
\end{figure*}

%Case when alpha fixed
To study the effect of maneuverability parameter ($\beta_0$) on the robot swarm, we first fixed $\alpha_0$ to a value sufficiently large for agents to modulate their speed in the presence of others. We then varied $\beta_0$ between 0, where agents do not modulate their turning rate in response to others, to 4, where agents react with a high degree of turning modulation to visual cues (see Fig.~\ref{fig:fig_robot_exp}A). We found that with sufficiently large parameter~$\beta_0$ groups achieved highly polarized movement with reduced temporal variance and inter-individual distance as they could sufficiently modulate their heading angle to remain in a single polarized group for extended periods. In this parameter regime group fragmentation was also less common, though still possible due to a limited FOV. The resulting collective motion was collisionless or collision avoidance events were rarely triggered by other robots due to the restricting effects of the arena walls. 

As expected, when $\beta_0$ was too low, agents could not modulate their turning rate enough to achieve high polarization or cohesive movement captured by low polarization and cluster size values. Furthermore, robots in this regime collided more often as they could not adjust their paths fast enough to avoid dangerously low inter-individual distances from group members increasing the risk of collisions. 

Conversely, when $\beta_0$ was too high, polarization also decreased compared to mid-range $\beta_0$ values due to an observed over-reactive turning behavior. In these regimes agents over-react to visible robots to the extent that they may easily lose sight of others resulting in group fragmentation. This is characterized by lower polarization and polarized cluster size and higher inter-individual distances due to diverging subgroups. 

Interestingly, although we found the same trend in all metrics with increasing $\beta_0$ values in simulations and robots, we found a mismatch between the optimal $\beta_0$ value between the two systems: in the robot swarm significantly larger $\beta_0$ values yield optimal performance compared to simulations. This mismatch might be caused by key differences resulting from the embodiment of agents in the collective. Next to other factors arising from friction and body weight, physical robots simply can not turn with arbitrary turning rates the same way as simulated agents can. Such differences highlight the importance of validating model results on embodied systems to assess the practicability of the model in real-world scenarios.

\FloatBarrier

\FloatBarrier

%Case when beta fixed
To study the effect of the velocity modulation parameter $\alpha_0$ we then fixed $\beta_0$ to a value large enough for robots to sufficiently modulate their turning rate. We changed $\alpha_0$ between 0, where agents do not modulate their speed, and 2, where agents react with large acceleration/deceleration to other visible robots (see Fig.~\ref{fig:fig_robot_exp}B).

Aligning well to our previous results with changing $\beta_0$, changing $\alpha_0$ revealed similar insights. With minimal $\alpha_0$ values robots couldn't sufficiently accelerate to remain cohesive and polarized and they could not decelerate to avoid physical collisions with each other. Increasing the parameter to an optimal mid-value showed emergent collective motion that is highly polarized, cohesive with little to no collisions avoidance events. Too large $\alpha_0$ values introduce over-reactive acceleration behavior and oscillations between forward and backward movement. Note that the vision-based model allows deceleration to negative velocities, i.e. robots can move backwards with respect to their heading, while their visual FOV is oriented forward. This might increases the chance of robot collisions with large $\alpha_0$ values as agents move faster and more frequently in a direction from which they can not gather visual information.

%Showcasing # TO DO VIDEO!
We demonstrate the behavior of the vision-based swarm by four representative videos\cite{mezey2024video4k}. First, we select one of the best achieving parameter combination, where both $\alpha_0$ and $\beta_0$ are from the optimal middle parameter ranges. We then recorded the resulting collective movement and collected the camera stream of agents and compiled these into a single video file (see Video~1~\cite{mezey2024video1}). During this experiment robots aligned with the arena wall and moved as a cohesive polarized group. We did not observe robot collisions or group fragmentation, although these both are possible due to a limited FOV and a confined space robots have to navigate in. We observed rare false positive detections (see Video~1~\cite{mezey2024video1}, Robot~1 from experiment time 01:03:30 to 01:05:00) of other visible objects outside the arena (such as visible curtains or occasional light reflections) mistakenly identified as robots that temporarily biased the resulting robot movement--an example of type of noise that can not fully taken into consideration with physics-based simulations. False negative detections of other robots were also rare but possible when others were too close to a focal robot (See Video~1~\cite{mezey2024video1}, Robot~2, from experiment time 11:52:00 to 12:21:00), temporarily increasing the probability of robot collisions.

Due to the limited FOV of robots, they can break away from the group (fission). But when these robots detect others again, they will rejoin the group (fusion). Although with a limited FOV rejoining events are more challenging than with a full FOV, we demonstrate that this is possible with the minimal vision-based model we use even with a half FOV. To present group fragmentation and rejoining events we recorded two videos. Video~2~\cite{mezey2024video2} shows a single robot breaking away from the swarm due to the the group's direction when arrived to the wall. As the group was almost orthogonal to the wall, the chance of some agents turning away from others was high. Once the robot, having previously split from the others, can see the group again, it can approach and rejoin it. Furthermore, in Video~3~\cite{mezey2024video3} we show similar joining behavior with two agents first breaking away from the group forming their own subgroup until others become visible again.

One of the main advantages of decentralized design is that no central data management or path calculation is necessary to achieve global order. As a result robots can be dynamically added to or removed from the group without disturbing the collective behavior. We show how the group can be extended on the fly by adding robots to the group one by one on Video~4~\cite{mezey2024video4}. Newly introduced robots speed up and join to the group ahead once others are in-sight. Removing robots on the fly is similarly unproblematic. Doing so, the group quickly recovers from missing members. When a robot is removed, remaining robots will modulate their speed to establish the preferred distance from others, keeping the collective movement intact.
With this we showcase stable and flexible flocking behavior of robots achieved through minimal, purely visual interactions and we underscore the promising practicality of the presented algorithm.  

\section*{Discussion}\label{discussion} % 1748 words
%Describing our results
In this work, we implemented a purely vision-based model of collective movement on a swarm of ten terrestrial robots with a limited FOV operating in confined spaces. Our preliminary analysis with agent-based simulations further supports that the model is resilient to perceptual and environmental constraints and can facilitate polarized, cohesive, and collisionless collective movement in simulated agents and robots. Our robot swarm is fully decentralized, with robots acting solely based on their camera streams without using central information, sharing information directly, or having memory of past events. Using this minimal vision-based controller, the swarm achieved highly polarized collective motion and maintained group cohesion. Most efficient collective motion was obtained especially when robots could modulate their speed and turning rate sufficiently to react but not overly react to others. These results show that our purely vision-based and decentralized robot control approach is feasible and effective in achieving robust collective movement of embodied agents.

%Discussing problems with FOV, i.e. losing others
Robots had a limited FOV, which could eventually lead to losing sight of each other and breaking away from the swarm causing group fragmentation especially in confined arenas. Our simulation results also support our observations in physical robots. Once the group is fragmented, it is more challenging to fuse into a single collective with a limited FOV. Future work could improve practical applicability by implementing a full, omnidirectional vision for robots, allowing fully collisionless and cohesive movement even in strictly limited spaces. As highlighted by our simulation results, such an improvement would also enable robots to show complex movement patterns, such as swarming or milling around a target point.

%Discussing occlusions and neighbor selection
Along with a limited FOV, visual occlusion shapes information transfer in visual networks \citep{poel2021spatial}. As a result, visual occlusion has a crucial impact on collective behavior arising from visual interactions. These include the flocking behavior of pedestrian crowds \citep{dachner2022visual}, the spatio-temporal characteristics \citep{kunz2012simulations} and collective detection capabilities of fish schools \citep{davidson2021collective}, social information use in collective foraging \citep{mezey2024visual}, or the performance of drone swarms using zonal flocking models \citep{schilling2022scalability}, among others. In the simplest case, without strict identification of peers, the input of the vision-based model implicitly accounts for visual occlusion, allowing overlapping projection “blobs” when robots occlude each other. In this case, the model considers the visual representation of each visible peer to construct the final binary input. On the other hand, using detection methods such as CNN-based object detection allows assigning individual robots to their bounding boxes on the registered camera image. This enables selecting only a few robots to interact with and naturally raises the question of neighbor selection: which visible neighbors should be considered during collective motion? Similar to fish paying attention only to a few visible peers \citep{jiang2017identifying, lei2020computational}, or to those that are the fastest~\citep{puy2024selective}, selecting which in-sight neighbors to follow might largely influence the vision-based model presented in this paper.

%Discussing problems with false detections (positive or negative) and slow detection rates
Robots in our framework detected others utilizing their camera streams, with an onboard CNN-based object detection network. Although recent advancements in edge computing allow near real-time inference with neural networks on edge devices, achieving practical usability for real-time motion control remains a challenge. The slower the detection of others, the slower the speed at which robots can reliably operate. Furthermore, false detections can influence the model by introducing noise to the visual input, resulting in movement perturbations that can only be partially modeled in physics-based robot simulators. While false positive detections can erroneously attract robots to unwanted targets, false negative detections might increase the probability that robots collide. Therefore, reducing the false detection rate and increasing inference speed are crucial in practical vision-based applications relying on such detection methods. In future work, improved object detection techniques could improve the visual input of the model leading to faster, more reliable, and robust collective movement in ever-changing environments. 

A possible approach to improve detection accuracy is to increase the input resolution of the camera stream. However, this would drastically decrease inference speed with the current technology. Another option is to extend the current system with event-cameras, which register slight changes in brightness with high temporal resolution and low resource usage. Combining event-based movement detection with traditional CNN-based techniques can overcome the drawbacks of frame-based methods and has already proven useful for long-term, high-resolution monitoring of penguin behavior \citep{hamann2023low}. Such a composite system could provide a truly real-time movement control while maintaining robot-detection accuracy.

Our results also generalize to systems capable of producing similar projection fields around focal robots as what we presented, including advanced segmentation techniques or selective LiDAR mapping. The primary challenge remains developing a robot-specific detection system that effectively filters out objects irrelevant to collective movement. In shoals of fish, for instance, social recognition of conspecifics is based on a wide range of cues depending on the species. These can be olfactory or acoustic signals, electric “fingerprints” in weakly electric fish, or general visual cues such as coloration, body shape, or even UV markings \citep{saverino2008social, griffiths2011social, ward2020social}.

%Discussing transferring results to aerial robotics
In our study, we used the 2-dimensional version of the purely vision-based model presented by \citet{bastien2020model}, which allowed us to provide vision-based decentralized control for terrestrial robots moving on a plane. Transferring our results to aerial robotics, implementing 3-dimensional motion modulation for unmanned aerial vehicles (UAVs), is an interesting future direction.

% Discussing scaling to larger groups and heterogeneous groups
Regarding scalability, a key question is how our results generalize to large robot swarms. As the number of robots and group density increases, the likelihood of collisions inherently rises, too. Here, the question of neighbor selection is especially crucial as the visual fields of robots become saturated in large groups. Therefore, generating reliable real-time input is key to achieve collisionless movement also in large groups.

% Discussing the importance of optic flow
The vision-based model we used integrates simple visual cues over the visual field of the focal agent to determine the direction and velocity of the agent's motion. Although \citet{bastien2020model} propose to optionally include optic flow in the state update rule, the effect of including such temporal derivatives of the visual cues is yet to be explored with this model. At the same time, on the other hand, it was shown in simulated particles that optic flow alone can account for similar patterns of collective motion~\citep{castro2024modeling} as those yielded by minimal visual cues presented in this paper. It is an interesting future direction to study how the combination of simple visual cues and optic flow co-shape collective behavior. Especially so, as the exact neural encoding of optic flow and its impact on collective animal behavior is still not clear~\citep{cronin2014visual, horrocks2023walking}. Taking a biorobotic perspective \citep{franceschini2014small, serres2017optic, horsevad2022beyond, lei2020computational}, the impact of optic flow on collective behavior could be systematically mapped in both robots and animals. In this context, mobile robot swarms would serve as embodied models of collective motion, to study the combined impact of optic flow and simple visual cues in collective systems. This approach would allow for the rigorous identification of scenarios where optic flow enhances or stabilizes collective performance. Forming a loop between natural and artificial systems, these findings could then generate hypotheses for further research in biology to better understand animal collectives and improve bio-inspired, vision-based robot swarms. On the other hand, although there are bio-inspired optic flow-based sensors, robot guidance methods~\citep{franceschini2004visual} and collective movement models~\citep{castro2024modeling, krongauz2024vision}, optic flow-based coordination of a collective in embodied swarm robot systems remains an unprecedented challenge.

% Mixed swarms
Our framework allows to introduce heterogeneity in collectively moving swarms. Firstly, being the presented model purely vision-based, if robots could identify a variety of distinct entities---from other robots or inanimate objects to animals, including humans---they could adapt their movement without the necessity for direct communication. By that, one could engineer composite movement strategies within mixed-entity swarms and hybrid societies~\citep{hamann2016hybrid} where direct communication is impossible, either due to different communication protocols (for example, differing robot-to-robot communication standards) or the complete absence of standard communication channels (e.g. between robots and animals). Such an approach has great potential in many applications~\citep{de2001animal}. It could help us designing adaptive robot swarms behaving according to the perceived human user's needs~\citep{kaduk2023effects} during human-swarm interaction~\citep{hasbach2022design}, or in wildlife rescue, pest control, in understanding animal collectives~\citep{landgraf2021animal} or improving robot swarms.

Secondly, as the vision-based model and every resulting application use a spatial scale defined by the body size of conspecifics, it is possible to flexibly introduce agents of different sizes while keeping the collective movement of the group intact---fundamentally different from classical models of collective motion. As an agent gets larger, its visual projection “blob” on others' retinas also grows, making others stay farther away. On the other hand, the stability of such mixed-sized robot swarms is yet to be explored. Furthermore, in its simplest form, the vision-based model does not distinguish between peers far from the focal agent and peers that are smaller than the focal agent. In future research, addressing this size-distance ambiguity in practical settings is key to adapting equilibrium distances for stable collective movement, for example by mapping detectable objects to their typical physical sizes.

Integrating computer vision datasets (e.g.\citenum{lin2014microsoft}) with various labelled objects can further enhance adaptability. By mapping detected objects to model parameters, robots could interact with a wide range of inanimate objects through avoidance or following behaviors. Additionally, a dictionary of observed objects would enable robots to dynamically adjust their movement parameters based on their surroundings. For instance, following an object might be beneficial in some contexts but not in others. This context-specific bias in movement patterns would allow for nuanced navigation in cluttered or dynamically changing environments. This approach could be extended with machine learning techniques to optimize collective movement according to arbitrary goals by tuning the parameter mapping. Using the vision-based model with such contextual tuning opens new directions for swarms operating in real-world scenarios.

% End
Our results offer promising future perspectives in vision-based swarm robotics. By decentralizing the control of individual robots, swarms can operate in challenging environments where centralized solutions break down either because of the absence of central knowledge or because of communication failures. With further refinements, our results have the potential to generalize to other input techniques and robot (or other agent) types in dynamic environments paving the way for diverse swarms including many, fundamentally different physical entities.

%TC:ignore
%\nolinenumbers
\onecolumn
\section*{Acknowledgements}\label{acknowled}
\subsection*{Contribution Statement}\label{contrib}
The initial concept was conceived by DM, RB, HH, and PR. 
The first draft was written by DM. All authors reviewed and contributed to the final version. 
DM implemented the agent-based simulations, analyzed the corresponding results, built the robot platform, and developed the vision-based robot controller. Supplementary videos, code, and data were deposited by DM, who also produced the supplementary videos and their components, as well as the main figures. RB created supplementary figure S1.
All these materials were reviewed by all coauthors.
Preliminary experiments for data collection were carried out by DM and YZ. 
The experimental infrastructure was prepared for final data collection, and the robot experiments were collaboratively planned and conducted by DM, DS, and YZ. 
The analysis pipeline for collected robot data was established and corresponding results were analyzed by DM and NM. NM deposited the corresponding code.

\subsection*{Code and Data Accessibility}\label{data-acc}
All relevant code and datasets used for our findings have been deposited: See \citenum{mezey2024simdata} for agent-based simulation framework and simulation data, \citenum{mezey2024robotdata} for experimental data, \citenum{mckee2024swarmvizdponce} for replay and analysis tool for experimental data and \citenum{mezeyVSWRM} for the vision-based robot controller. Supplementary Videos are available on the TIB-AV portal of the Leibniz Information Centre for Science and Technology University Library as a video series~\cite{mezey2024video1, mezey2024video2, mezey2024video3, mezey2024video4} and via Zenodo in 4K resolution\cite{mezey2024video4k}.

\subsection*{Disclosure Statement}\label{disclosure}
During the preparation of this work, DM used the large language model (LLM) GPT-4/4o to improve various parts of the text for (1) readability, flow and clarity, (2) grammatical correctness, and (3) simplicity. The LLM was prompted to rate snippets of the draft according to a list of criteria corresponding to these three points and to provide detailed reasoning for the ratings. Grammatical improvements were then added to the text, after which single synonyms, alternative word compositions, and in some cases sentence-level reformulations were introduced. No changes other than stylistic or grammatical were introduced during this process.
After using this tool, all authors reviewed and edited the content as needed and take full responsibility for the content of the publication.

\subsection*{Competing Interests}
The authors declare no competing interests.

\subsection*{Funding}\label{funding}
Funded by the Deutsche Forschungsgemeinschaft (DFG, German Research Foundation) under Germany’s Excellence Strategy – EXC 2002/1 “Science of Intelligence” – project number 390523135.

\section*{Supplementary Notes}
\subsection*{Calculating Visual Social Interactions}\label{sup_examples}
Social interactions are calculated from (1) the area of visual projection ``blobs" ($V_i(\phi_i, t)$) and (2) their edges ($(\frac{\partial { V_i(\phi_i, t)}}{\partial \phi_i})^2$). These are modulated by a trigonometric mask allowing differentiating between (1) front and back ($\cos(\phi_i)$) or (2) between left and right ($\sin(\phi_i)$). These trigonometric masks model the symmetry of the agents' visuo-motor system. For instance, an attraction force in the turning response of the focal agent (turning towards a conspecific) will depend on which side the focal agent perceives the corresponding visual blob. The same attraction force shall yield turning right when the other agent is on the right but turning left when the other agent is on the left. Similarly, an attraction force in the focal agent's velocity (accelerating towards a conspecific) shall yield acceleration when the other agent is in front of, but deceleration when the other agent is behind the focal agent. Note that any odd function (changing sign at 0) would suffice to model left-right symmetry and any even function (changing sign at $\pm\frac{\pi}{2}$) to model front-back symmetry of this kind.

The modulated area term (e.g. $\int\cos{\phi}V_i(\phi, t)d\phi$) increases with larger visual blobs i.e. closer visible agents, hence it corresponds to short range interactions. The modulated edge term (e.g. $\int\cos{\phi}\frac{V_i(\phi, t)}{d\phi}d\phi$) increases with narrower blobs, i.e. farther visible agents, hence it corresponds to long-range interactions. Summed with their opposite signs these terms generate long-range attraction and short-range repulsion forces on agents purely arising from visual perception (see Fig.~\ref{fig:fig1}C). 

To give a detailed example for acceleration response we refer to \citet{bastien2020model} Fig.~2 and Fig.~\ref{fig:fig_s1}: When a another agent is far in front of the focal agent it's corresponding visual projection blob is central and narrow. As a result, its' small overlapping area with the cosine mask causes a small short-range repulsion. At the same time its' high edges near to each other around $\phi_i=0$ where $\cos(\phi_i)$ has a maximum, cause a large positive long-range attraction. Summed together these two cause an overall positive change in the agent's velocity (acceleration). In contrast, when another agent is too close in front of the focal one, the corresponding visual blob is wide. Therefore when modulated with $\cos(\phi)$, edges become lower as they get more peripheral. This results in smaller long-range attraction than in the previous case. Due to the large modulated area of the blob now a large short-range repulsion arises. The sum of these two will result in an overall negative change in velocity (deceleration). 

Similarly, the turning rate of the agent is modulated via long-range attraction and short-range repulsion forces calculated from projection blob area and edge size. Note, that turning behavior is shaped by the $\sin$ mask instead of $\cos$ that was used for acceleration. When an agents is far ahead, the focal agent turns towards it according to the overall attraction coming from the corresponding small blob area and it's close edges. On the other hand, when the visible individual is near, the focal agent turns away due to the overall repulsion coming from the large blob area and small modulated edges.

\renewcommand{\thefigure}{S1}
\begin{figure*}%[h]
\centering
\includegraphics[width=0.95\textwidth]{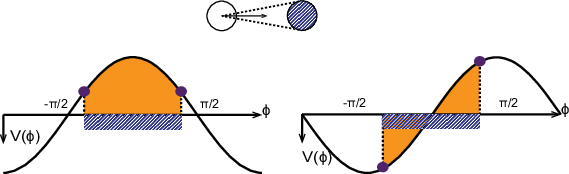}
\caption{\label{fig:fig_s1} \textbf{Calculation of social acceleration and turning response} The visual input of a focal agent (white disc) with another agent directly ahead (striped disc) is modulated through \textit{cos} (left) and \textit{sin} (right) masks to calculate the social acceleration and turning response respectively. The modualted area terms (i.e. $\int\cos{\phi}V_i(\phi, t)d\phi$ and  $\int\sin{\phi}V_i(\phi, t)d\phi$) are shown with orange highlight while the modulated edge terms (i.e. $\int\cos{\phi}\frac{V_i(\phi, t)}{d\phi}d\phi$ and $\int\sin{\phi}\frac{V_i(\phi, t)}{d\phi}d\phi$) with dotted lines.}
\end{figure*}

\subsection*{Limited FOV}\label{sup_limfov}
When limiting agents' FOV, partial blobs might occur on the peripheries of the active visual field (i.e. where $|\phi| \leq \phi_L$). Note that even if a partial projection blob is small --- and small visual blobs correspond to far away agents --- it might actually correspond to a nearby agent that would yield large blobs when fully visible otherwise (See also Fig.~\ref{fig:fig_FOV_patterns}A). This mismatch between partial detection blob size and distance information might perturb collective movement as it introduces unwanted attraction on the peripheries of the visual field (boundary effects). To overcome such corrupted sensory information we used heuristics in both simulations and in robots.
In simulations, when a detection blob is only partially in the active visual field (i.e. $V_i(\phi_L+1, t)=1$ or $V_i(-\phi_L-1, t)=1$), we assume that it's a partial projection blob and we recover the whole blob even if that extends over the limits of the active vision. 
In robot agents we use a similar method. Here, we have a 2D visual perception where bounding boxes have not only widths, but heights as well, and we can use this information to recover the full detection box by assuming detection boxes of other robots to be square. If this is not the case and the detection box touches the left- or rightmost pixel of the robot's camera stream, we assume that it is a partial detection and we extend the resulting projection blob according to the height of the detection box instead of using only its partial width.
Our heuristic recovery of peripheral blobs is based on the assumption that partially seeing a nearby conspecific can be cognitively distinguished from fully seeing a far away conspecific in natural agents. This is a reasonable assumption, if agents can detect and recognize each other as conspecifics, or they have an internal model about the typical size of group members.

\subsection*{Simulations On a Torus}\label{sup_torus}
To allow direct comparison of collective movement of simulated groups with limited FOV with those presented by~\citet{bastien2020model} we used an arena with periodic boundaries. The width and height of the arena was set to 900~pixels, which is approximately 164 times the radius agents ($R_A=5.5$ pixels). An arena with periodic boundaries allows agents leaving on one side to reappear instantaneously on the other side. This solution is equivalent to simulating collective movement on the surface of a torus. Simulating collective movement in such an arena mimics the properties of an unlimited environment when the group is cohesive. When due to a limited FOV the group fractures, on the other hand, agents previously broken away will reappear in the arena allowing fusion of the group, that an unlimited environment would not support resulting in catastrophic group fragmentation.

Visual interactions in such a toroidal environment can carry some ambiguities, as with an unlimited visual range a focal agent could theoretically have access to multiple visual representations of the same agent. For example if a visible agent is straight ahead of the focal one, due to the periodic boundary conditions the visual representation of this agent could be accessed both from a relative direction of 0 (from the front) and from a relative angle of $\pm\pi$ (from the back through the periodic boundaries). In other words, multiple visual ``copies" of the same agent would be accessible in different relative directions and to different distances.

To account for such ambiguities on a torus, we always take only a single visual representation of a visible agent into account when calculating the visual input of the focal individual. To choose which ``copy" of the visible agent shall be used to create the visual projection field of the focal agent, we use a simple tiling method of the environment. First we generate eight copies of the visible agent $i$ with position $(x_i, y_i)$ by adding or subtracting the arena width ($W$) or height ($H$) from its first and second coordinates. This yields a set of agent copies $C_i$ including the original visible agent coordinates as follows:
\begin{equation}
    C_i = \{ \vec{x}_j = (x_j, y_j)\ |\ x_j\in[x_i-W, x_i, x_i+W],\ y_j\in[y_i-H, y_i, y_i+H] \}
\end{equation}

From these we then choose the agent copy with position $\vec{x}_k$ that minimizes euclidean distance to the focal agent, i.e.
\begin{equation}
    \vec{x}_k \in C_i\ |\ k = \argmin_{j}\sqrt{(x_j-x_i)^2+(y_j-y_i)^2}
\end{equation} 

The visual projection corresponding to the selected copy is then added to the visual projection field of the focal agent. 

\subsection*{Metrics}\label{sup_metrics}
During this study we used five key metrics: group polarization ($P$), mean inter-individual distance ($\tilde{D}$), the ratio of simulation/experiment time agents spent in overlap or in collision avoidance mode for actual robots ($R_o$), group circularity ($RCA$), and group fragmentation through hierarchical clustering of the agents.

Being $\vec{x}_i(t)$ the position vector and $\vec{n}_i(t)$ the unit heading vector of agent $i$ at time t and, $N_A$ is the total number of agents, $\vec{h}_i(t)$ is the position vector of the vertex $i$ of the convex hull surrounding the group (ordered by adjacency) with $N_h$ vertices in total the above metrics are formalized as follows:

Polarization order ($P$) is defined as the length of the mean of the agents' unit heading vectors and it indicates the uniformity of movement direction within the group. It reaches high values close to 1 when all agents move in similar directions while low values close to zero for either unorganized movements where all agents move in different directions or when the group is fragmented into subgroups. 

%Equations
\begin{equation}
    P(t) = \frac{1}{N_A} \left| \left| \sum_{i=1}^{N_A} \vec{n}_i(t) \right| \right|
\end{equation}

The mean inter-individual distance ($D$) is defined as the average distance between agents.

\begin{equation}
    \tilde{D}(t) = \frac{2}{N_A(N_A-1)}\sum_{i=1}^{N_A}\sum_{0 < j < i} \left\| \vec{x}_i(t) - \vec{x}_j(t) \right\|
\end{equation}

Given the diameter of the surrounding convex hull defined by the agents in the group (i.e. the distance between the farthest two non-adjacent vertices):

\begin{equation}
    d(t) = 
    \begin{cases}
    \max_{i, j}  \left| \left| \vec{h}_i(t) - \vec{h}_j(t)\right| \right| \quad \text{such that } 1<|i-j|<(N_A-1) \quad &\text{if } N_h > 3 \\
    \max_{i, j}  \left| \left| \vec{h}_i(t) - \vec{h}_j(t)\right| \right| \quad &\text{if } N_h \leq 3
    \end{cases}   
\end{equation}

And using the circular convention of $\vec{h}_{N_h+1}=\vec{h}_{1}$, the convex hull's area as:

\begin{equation}
    A(t) = \frac{1}{2} \sum_{i} \left( h_{i,1}(t) h_{i+1,2}(t) - h_{i+1, 1}(t) h_{i,2}(t) \right)
\end{equation}

The circularity of groups ($RCA$) can be defined by comparing the convex hull's area with that of a circle with the same diameter ($A_C(d(t))$). An $RCA$ value close to 1 signifies a perfectly circular group, whereas values approaching 0 indicate elongated formations.

\begin{equation}
    RCA(t) = \frac{A(t)}{A_C(d(t))} = \frac{A(t)d(t)^2\pi}{4}
\end{equation}

In simulated agents $R_o^{sim}$ shows what proportion of the average simulation time (in percent) agents spent overlapping and is directly related to agent-agent collision probability in embodied groups. Consequently, $R_o^{sim}$ reaches 100 if all agents overlapped through the whole simulation time and 0 if no agent-agent overlaps were observed.

$D_{ij}$ being the distance between agent $i$ and $j$ and $R_A$ the radius of agents, $R_o^{sim}$ can be formulated as follows:

\begin{equation}
    R_o^{sim} = \frac{1}{2N_A}\sum_{i=1}^{N_A}\frac{T_o}{T}100 = \frac{1}{2N_A}\sum_{i=1}^{N_A}\frac{\#\{t: \exists j | D_{ij}(t)<2R_A, j\in\{1,...,N_A\}, j \neq i\}}{T}100
\end{equation}

In robot experiments overlaps are not possible, hence $R_o^{exp}$ is defined by the proportion of the total experiment time (in percent) robots spent in collision avoidance mode due to other robots. As a result, $R_o^{exp}$ reaches 100 if all robots spent the whole experiment in obstacle avoidance mode due to other robots and 0 if no robot avoidance protocol has been activated at all.

$C_i(t)=1$ denoting robot $i$ being in robot avoidance mode at time $t$ while this value being 0 denoting normal operation of robot $i$, $R_o^{exp}$ can be formulated as:

\begin{equation}
    R_o^{exp} = \frac{1}{N_A}\sum_{i=1}^{N_A}\frac{T_c}{T}100 = \frac{1}{N_A}\sum_{i=1}^{N_A} \frac{\#\{t: C_i(t)=1\}}{T}100
\end{equation}

Furthermore, we clustered agents according to their movement similarity (i.e. similarity based on their heading angle and distance from others) to quantify group cohesion. This way we could define polarized and cohesive subgroups even if these were otherwise not forming a single group. This is possible, as agents' can lose sight of each other due to their limited FOV leading to multiple leaders and subgroups following them. We then calculated the size of the largest polarized, cohesive subgroup $N_{clus}^{max}$. This is $N_{clus}^{max}=N_A=10$ if all agents moving together in close vicinity towards the same direction and 1 if each agent is relatively far from others while they move in different directions. We chose the threshold of the clustering method heuristically as \textit{0.275}, such that it yields clustering comparable with our judgment for polarized subgroups.

For hierarchical clustering in simulations we used our custom simulation framework \citep{mezey_P34ABM_A_novel} and the \textit{fastcluster} \citep{müllerfastcluster} python package with Ward clustering scheme, with dissimilarity metric defined as follows. Given the inter-individual distance matrix in time $t$ as

\begin{equation}
    \Bar{\Bar{\mathbf{D}}}(t) : D_{i,j}(t) = || \vec{x}_i(t) - \vec{x}_j(t) ||
\end{equation}

The normalized distance matrix is defined as 

\begin{equation}
    \Bar{\Bar{\mathbf{D^N}}}(t): D^N_{i, j}(t) = \frac{| \median(\Bar{\Bar{\mathbf{D}}}(t))-D_{i,j}(t) |}{\max(\Bar{\Bar{\mathbf{D}}}(t))}
\end{equation}

And the final dissimilarity matrix is defined as:

\begin{equation}
    \Bar{\Bar{\mathbf{M}}}(t): M_{i,j}(t) = \frac{(1 - P_{i,j}(t) + D^N_{i,j}(t)}{2}
\end{equation}

where $P_{i,j}(t)$ is the polarization value between agent $i$ and $j$ in time $t$, i.e.

\begin{equation}
    P_{i,j}(t) = \frac{||\vec{n}_i(t)+\vec{n}_j(t)||}{2}
\end{equation}

Similarly, for robot data we used hierarchical clustering taking the robots' orientation and distance into consideration using our custom visualization and analysis tool developed in Julia \citep{ mckeeswarmvizgithub, mckee2024swarmvizdponce}. We used the clustering threshold of \textit{0.1653} and dissimilarity matrix:

\begin{equation}
    M_{i,j}(t) = 1 - \sqrt{\left(1 - \frac{\left| \left| \vec{x}_i(t) - \vec{x}_j(t) \right| \right| }{r_{\text{max}}}\right) \frac{\vec{n}_i(t) \cdot \vec{n}_j(t) +1}{2}}
\end{equation}

where

\begin{equation}
    r_{max} = \max_{i,j,k,l,t_*} \sqrt{\left( x_{i,1}(t_1)-x_{j,1}(t_2) \right)^2 + \left( x_{k,2}(t_3)-x_{l,2}(t_4) \right)^2}
\end{equation}

\subsection*{Collision Avoidance}\label{sup_collisions}
During experiments, depending on the selected model parameters, it can happen that robots move too close to each other. To avoid possible robot-robot collisions and to reflect robots from arena walls we implemented a custom obstacle avoidance process running parallel with the robot controller. When an obstacle ahead is detected with one of the five frontal proximity sensors of the Thymio~II base robot, a recursive emergency protocol takes over, stops the robot and rotates it to the direction where the obstacle signal is lower (turning away from the obstacle) until the path ahead is clear. Symmetric obstacles are avoided with a fixed-direction, 90 degree turn. In case the proximity sensors in the back of the robots are activated, robots stop and retry to act according to their vision-based controller in every two seconds. This is practical when robots would go backwards but the arena walls do not allow them. Robot collision avoidance events are not recorded, but given the fixed, relatively high rotational speed of the robots during collision avoidance, it is easily detectable from robot tracking data using the angular velocity of robots and their distance from other robots or from walls.

\subsection*{From Detection to Robot Movement}\label{sup_detectiontomovement}
To generate model input from the agents' raw camera streams robots have to detect each other in their environment. In our framework, robots use the ``SSD MobileNet V2 FPNLite 320$\times$320" object detector, a freely available convolutional neural network-based solution \citep{huang2017speed, tf2modelzoo} with a feature-pyramid (FPN) feature extractor network \citep{lin2017feature}. Such an object detector can provide bounding box coordinates of detected robots on the camera frames after training. We trained the network with ca. 2000 original images taken by robots in realistic scenarios using tensorflow 2\citep{developers2022tensorflow}. The dataset was labeled and augmented using Roboflow \citep{dwyer2022roboflow}. The trained model was full-integer quantized and compiled such that inference could be delegated to Google Coral Edge TPU USB accelerator devices \citep{coraledgetpu} onboard. This enabled a stable 5 Hz object detection rate on the edge devices with a 320 by 200 pixel (stretched to 320 $\times$ 320 pixel) image resolution during experiments. Bounding boxes that overlapped more than 50\% or that was smaller than 3 pixels wide have been filtered during experiments. After detecting robots within the FOV, the resulting bounding boxes are projected to the horizontal axis to produce a visual projection field (See Fig.~\ref{fig:fig_robotplatform}B, steps 2-3). As attraction-repulsion forces depend on the width of the resulting blobs and blobs are truncated on the peripheries due to a limited FOV, we recovered partial peripheral detections as described in section \nameref{sup_limfov} to avoid boundary effects. The resulting visual projection field is transformed to a pair of theoretical acceleration and turning rate values using the vision-based model. The output is then translated to motor velocity values via a differential steering algorithm \citep{lucassteering}. The absolute motor velocity of robots has been limited to ca. 75\% of the maximum motor velocities of the ThymioII base robots.

\clearpage

%TC:endignore
\begin{table}
\small % Make the whole table smaller
\begin{tabularx}{\textwidth}{lXlll}
\hline
\rowcolor{lightred} Parameter & Description  & Value & Unit & Variable Name \\
\hline
$W$     & Arena width & 900 & px & \scriptsize{ENV\_WIDTH}           \\
\hline
$H$     & Arena height & 900 & px & \scriptsize{ENV\_HEIGHT}           \\
\hline
$N_{ret}$     & Resolution of the visual projection input of agents & 320 & px & \scriptsize{VISUAL\_FIELD\_RESOLUTION}           \\
\hline
$R_A$     & Half bodylength of agents & 5.5 & px & \scriptsize{RADIUS\_AGENT}           \\
\hline
$\gamma$     & Self-propulsion preference factor & 0.1 & - & \scriptsize{VF\_GAM}           \\
\hline
$v_0$     & Preferred individual speed & 1 & px/ts & \scriptsize{VF\_V0}           \\
\hline
$\alpha_1$     & Parameter controlling front-back equilibrium distance $L^{fb}_{eq}$ & 0.09 & - & \scriptsize{VF\_ALP1}            \\
\hline
$\beta_1$     & Parameter controlling left-right equilibrium distance $L^{lr}_{eq}$ & 0.09 & - & \scriptsize{VF\_BET1}            \\
\hline
$\alpha_0$     & Parameter controlling social acceleration/deceleration & var & - & \scriptsize{VF\_ALP0}            \\
\hline
$\beta_0$     & Parameter controlling social turning response & var & - & \scriptsize{VF\_BET0}            \\
\hline
$2\phi_L$     & Field of view & var & - & \scriptsize{AGENT\_FOV}           \\
\hline
$T$     & Simulation time & 20000 & ts & \scriptsize{T}           \\
\hline
$N$     & Number of agents & 10 & - & \scriptsize{N}           \\
\hline
-     & Arena type & torus/walls & - & \scriptsize{BOUNDARY}           \\
\hline
-     & Vision range & 2000 (i.e. $\inf$) & px & \scriptsize{VISION\_RANGE}           \\
\end{tabularx}
\caption{Fixed and varied (var) parameters of the simulation framework used for Fig.~\ref{fig:fig_FOV_patterns} and \ref{fig:fig_wall_metrics}. \textit{px} denotes pixels, and \textit{ts} simulation timesteps.}\label{tab_simulationparams}
\end{table}

\begin{table}
\small % Make the whole table smaller
\begin{tabularx}{\textwidth}{lXll}
\hline
\rowcolor{lightred} Parameter & Description  & Value & Unit \\
\hline
$W$     & Arena width & 900 & cm         \\
\hline
$H$     & Arena height & 600 & cm          \\
\hline
$N_{ret}$     & Resolution of the visual projection input of agents & 320 & px   \\
\hline
$R_A$     & Half bodylength of agents & 5.5 & cm         \\
\hline
$\gamma$     & Self-propulsion preference factor & 0.1 & -           \\
\hline
$v_0$     & Preferred individual speed & 90 & arbitrary motor unit       \\
\hline
$\alpha_1$     & Parameter controlling front-back equilibrium distance $L^{fb}_{eq}$ & 0.09 & - \\
\hline
$\beta_1$     & Parameter controlling left-right equilibrium distance $L^{lr}_{eq}$ & 0.09 & -         \\
\hline
$\alpha_0$     & Parameter controlling social acceleration/deceleration & var & -           \\
\hline
$\beta_0$     & Parameter controlling social turning response & var & -       \\
\hline
$2\phi_L$     & Field of view & ca. 175 & deg          \\
\hline
$T$     & Experiment time & ca. 60 & min          \\
\hline
$N$     & Number of agents & 10 & -         \\
\hline
-     & Arena type & walls & -       \\
\hline
-     & Vision range & 3-6 & m         \\
\end{tabularx}
\caption{Parameters used during robot experiments for Fig.~\ref{fig:fig_robot_exp}.}\label{tab_experiment}
\end{table}

\end{document}